\documentclass[pdf-a,balance,colorlinks,upint,subscriptcorrection,varvw,mathalfa=cal=boondoxo, spanish,french,vietnamese,russian,greek,grid,nofoot]{asmeconf}
\usepackage{comment}

\usepackage{setspace}
\usepackage{times}
\usepackage{graphicx}
\usepackage{url}
\usepackage{graphicx}
\usepackage{subcaption}
\usepackage{amsmath}
\usepackage{float}
\usepackage{multicol}
\usepackage{multirow}
\usepackage{datetime}
\newdateformat{customdate}{\ordinal{DAY} \monthname[\THEMONTH], \THEYEAR}
\usepackage{xcolor}
\usepackage{tcolorbox}

\definecolor{cbblue}{HTML}{0072B2}
\definecolor{cbgreen}{HTML}{009E73}
\definecolor{cborange}{HTML}{FF8000}
\definecolor{cbgray}{HTML}{7F7F7F}

\newtcolorbox{userpromptbox}{
    colframe=cbblue,
    colback=cbblue!5,
    fonttitle=\bfseries,
    boxrule=0.5mm,
    arc=3mm,
}

\newtcolorbox{chatgptresponsebox}{
    colframe=cbgreen,
    colback=cbgreen!5,
    fonttitle=\bfseries,
    boxrule=0.5mm,
    arc=3mm,
}

\definecolor{mydarkgreen}{RGB}{0,200,0}
\definecolor{mydarkred}{RGB}{200,0,0}
\usepackage{hyperref}
\hypersetup{
    colorlinks=true,    
    citecolor=blue,     
    linkcolor=blue,     
    urlcolor=blue       
}

\usepackage{graphicx}
\usepackage{bbding}
\usepackage{tabularx,colortbl}
\usepackage{lscape}
\usepackage{adjustbox}
\usepackage{pdflscape}
\usepackage{setspace,lipsum}
\usepackage{caption}
\usepackage{titlesec}
\titlespacing*{\section}{0pt}{*0.5}{*0.5} 
\titlespacing*{\subsection}{0pt}{*0.5}{*0.5} 
\titlespacing*{\subsubsection}{0pt}{*0.5}{*0.5} 
\usepackage{enumitem}
\setlist[itemize]{itemsep=5pt, topsep=2pt} 
\setlist[enumerate]{noitemsep, topsep=1pt} 

\usepackage{soul}

\hypersetup{%
	pdfauthor={Mohamed Elrefaie},									  
	pdftitle={AI Design Agents Elrefaie},                  
	pdfkeywords={ASME conference paper, LaTeX template, BibTeX style},
	pdfsubject = {Describes the asmeconf LaTeX template},			  
	pdflicenseurl={https://ctan.org/pkg/asmeconf},
}

\usepackage{soul}

\begin{document}









\title{\NoCaseChange{AI Agents in Engineering Design: A Multi-Agent Framework for Aesthetic and Aerodynamic Car Design\vspace{10pt}
}} 

\SetAuthors{%
	Mohamed Elrefaie\affil{1}\CorrespondingAuthor{mohamed.elrefaie@mit.edu}, 
	Janet Qian\affil{2}, 
        Raina Wu\affil{1}, 
        Qian Chen\affil{1},
	Angela Dai\affil{3}, 
	Faez Ahmed\affil{1}}

\SetAffiliation{1}{Department of Mechanical Engineering, Massachusetts Institute of Technology, Cambridge, MA, USA}
\SetAffiliation{2}{Department of Electrical Engineering and Computer Science, Massachusetts Institute of Technology, Cambridge, MA, USA}
\SetAffiliation{3}{Department of Computer Science, Technical University of Munich, Munich, Germany}


\maketitle

\versionfootnote{Preprint - Under Review}





\begin{abstract}
We introduce the concept of \textbf{``Design Agents''} for engineering applications, particularly focusing on the automotive design process, while emphasizing that our approach can be readily extended to other engineering and design domains. Our framework integrates AI-driven design agents into the traditional engineering workflow, demonstrating how these specialized computational agents interact seamlessly with engineers and designers to augment creativity, enhance efficiency, and significantly accelerate the overall design cycle. By automating and streamlining tasks traditionally performed manually, such as conceptual sketching, styling enhancements, 3D shape retrieval and generative modeling, computational fluid dynamics (CFD) meshing, and aerodynamic simulations, our approach reduces certain aspects of the conventional workflow from weeks and days down to minutes. These agents leverage state-of-the-art vision-language models (VLMs), large language models (LLMs), and geometric deep learning techniques, providing rapid iteration and comprehensive design exploration capabilities. We ground our methodology in industry-standard benchmarks, encompassing a wide variety of conventional automotive designs, and utilize high-fidelity aerodynamic simulations to ensure practical and applicable outcomes. Furthermore, we present design agents that can swiftly and accurately predict simulation outcomes, empowering engineers and designers to engage in more informed design optimization and exploration. This research underscores the transformative potential of integrating advanced generative AI techniques into complex engineering tasks, paving the way for broader adoption and innovation across multiple engineering disciplines.


\end{abstract}







\begin{figure*}[h!]
    \centering
    \includegraphics[width=\linewidth]{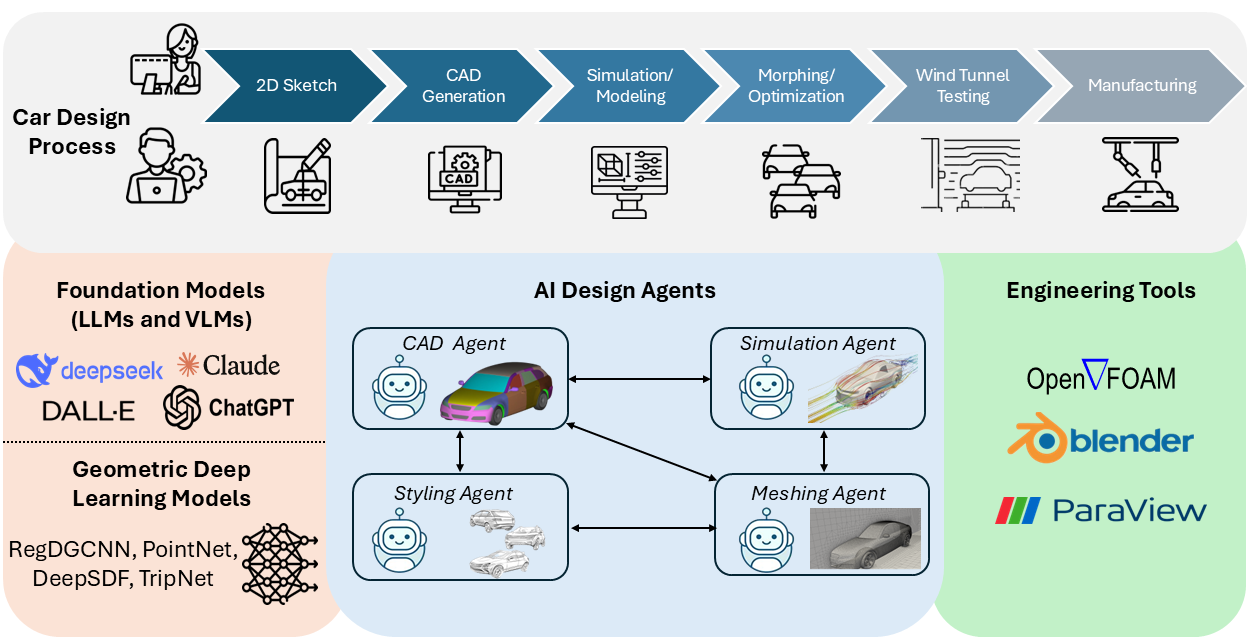}
\caption{\textnormal{AI Design Agents for Accelerating the Car Design Process. The framework integrates vision-language models (SDXL, ControlNet, DALL·E), geometric deep learning models (DeepSDF, PointNet, RegDGCNN, TripNet), and LLMs to automate design tasks. AutoGen helps different agents communicate with each other, improving coordination and making the design process faster and more efficient. The agents can also interact with various engineering tools and execute Python commands, enabling automation of complex design and simulation workflows.}}
\label{fig:ai_design_agents}
\end{figure*}

\section{Introduction}
\label{sec:Introduction}
The design of a car is a multidisciplinary endeavor that balances engineering performance with aesthetic appeal. Unlike aircraft design, where functional performance and safety take absolute priority, automotive design is deeply influenced by styling and brand identity, making it both an engineering and artistic process~\cite{forslund2010aesthetic, li2019research, li2019same, catalano2007framework, cheutet20072d, cheutet20053d, jagtap2015aesthetic, yadav2013prioritization, liu2017effects}. Consumers are drawn to cars not only for their technical specifications but also for their visual appeal, as aesthetics play a crucial role in market success~\cite{burnap2021design}. Consequently, car design involves a complex interplay between aerodynamics, manufacturability, and subjective user preferences. Traditional design workflows rely on iterative refinements where designers sketch conceptual ideas, evaluate existing designs for inspiration, and collaborate with engineers to assess performance and feasibility.


While traditional car design workflows often involve a slow and iterative process of sketching, evaluation, and engineering refinement, we propose a novel approach that integrates \textbf{AI Design Agents} into the conceptual design phase. In this context, agents are specialized, autonomous computational systems designed to perform specific design tasks by leveraging artificial intelligence, machine learning, and automation. These agents automate critical tasks, streamlining the transition from initial sketches to fully simulated aerodynamic evaluations. The multi-agent framework introduced in this work covers key stages of the car design pipeline, beginning with early-stage sketching and styling, incorporating large language models (LLMs) and vision-language models (VLMs) such as Stable Diffusion XL~\cite{podell2023sdxl} and ControlNet~\cite{zhang2023adding} to enhance and refine conceptual designs. The process continues with automated 3D shape retrieval and generation, where geometric deep learning models transform hand-drawn sketches into 3D car meshes or retrieve similar designs from the DrivAerNet++\footnote{\url{https://github.com/Mohamedelrefaie/DrivAerNet}}~\cite{elrefaie2024drivaernet, elrefaie2025drivaernet++}  database. Additionally, deep learning-based surrogate models allow for rapid aerodynamic evaluations, replacing expensive and time-consuming CFD simulations with predictive models trained on high-fidelity data. To ensure seamless integration into simulation workflows, an AI-driven meshing agent is employed to modify and adapt meshing parameters within OpenFOAM~\cite{OpenFOAMv11}, facilitating automated and optimized aerodynamic analyses.

This multi-agent framework not only accelerates the design cycle but also fosters a more collaborative interaction between designers and engineers. By leveraging geometric deep learning and nonlinear dimensionality reduction techniques, such as \textit{t-distributed stochastic neighbor embedding} (t-SNE)~\cite{van2008visualizing}, we can have a better understanding of the high-dimensional design space of DrivAerNet++. This approach enables the identification of clusters of high-performance designs, facilitates the analysis of critical design features, and provides a method for visually validating CAD/CAE results. Furthermore, the entire process is automated through Python APIs, utilizing AutoGen~\cite{wu2023autogen} for AI agents orchestration.

\begin{figure*}[h!]
    \centering
    \includegraphics[width=\linewidth]{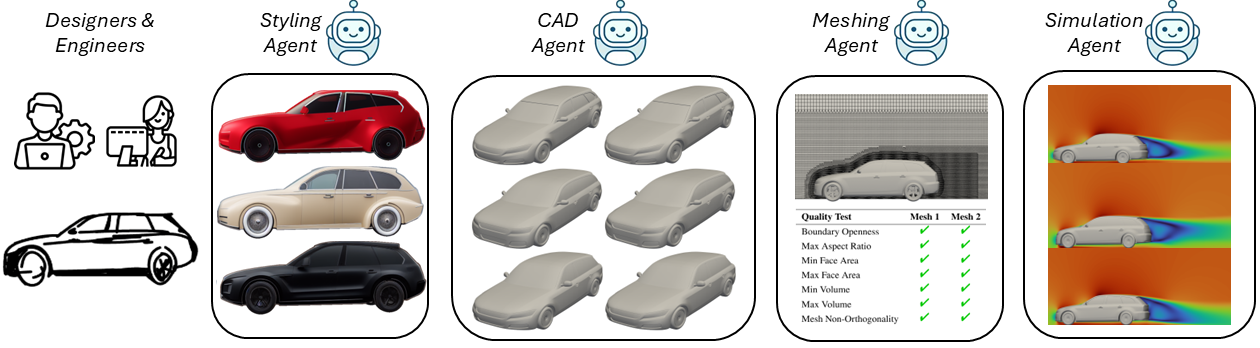}
\caption{Our multi-agent system enables effective interaction between engineers and designers throughout the car design process. Given an input 2D sketch for conceptual design, the \textbf{Styling Agent} renders high-resolution images, enhancing visual aesthetics. The \textbf{CAD Agent} generates new designs via generative modeling or retrieves 3D meshes from the DrivAerNet++ database. The \textbf{Meshing Agent} creates high-quality computational meshes for CFD simulations and evaluates mesh quality. Finally, the \textbf{Simulation Agent} performs real-time aerodynamic performance predictions or retrieves aerodynamic data from the DrivAerNet++ database, accelerating the iterative design process. Different outputs from the various agents can be used for design exploration or design optimization, enabling a data-driven and efficient approach to automotive design.}
    \label{fig:multi_agent_system}
\end{figure*}

Figure~\ref{fig:ai_design_agents} illustrates our AI-driven framework for accelerating the car design process, highlighting the iterative nature of collaboration between designers, stylists, and engineers. The conventional car design workflow begins with a 2D sketch or conceptual design, followed by CAD modeling and 3D geometry generation. Subsequent steps involve running physics-based simulations, such as computational fluid dynamics (CFD) and finite element analysis (FEA), before refining the design through morphing and optimization. The validated designs then proceed to wind tunnel testing and, ultimately, manufacturing.

On the other hand, our AI Design Agent framework integrates vision-language models (VLMs) and large language models (LLMs) alongside geometric deep learning methods, significantly accelerating and enhancing these traditional workflows. This integration enables seamless interaction with engineering software such as Blender~\cite{blender}, OpenFOAM~\cite{OpenFOAMv11}, and ParaView~\cite{ahrens200536}. The system consists of four specialized design agents:

\begin{itemize}
    \item  \textit{Styling Agent}, responsible for generating high-resolution renderings and enhancing aesthetic appeal;
    \item \textit{CAD Agent}, which facilitates design retrieval from the DrivAerNet++ dataset and enables generative 3D shape modeling based on industry-standard designs; 
    \item \textit{Meshing Agent}, which interacts with OpenFOAM to generate high-quality CFD meshes for aerodynamic simulations; and 
    \item \textit{Simulation Agent}, which leverages surrogate models to provide real-time aerodynamic predictions while also retrieving simulation results from the database. 
\end{itemize}

Our approach significantly accelerates the iterative design process, bridging the gap between conceptualization, performance evaluation, and optimization. Figure~\ref{fig:multi_agent_system} illustrates how engineers and designers interact with the multi-agent system, integrating conceptual sketches, styling renderings, CAD retrievals, mesh generation, and real-time aerodynamic simulations for streamlined automotive design.

This paper is organized as follows: Section~\ref{sec:related_work} reviews related work on the integration of generative AI tools in engineering design, with a particular focus on applications in car design. Section~\ref{sec:DesignAgents} presents the proposed methodology, detailing the implementation of various AI design agents and their roles in the conceptual design process. Section~\ref{sec:results} provides results and observations from our experiments, highlighting the efficiency and effectiveness of the proposed AI-driven framework. Section~\ref{sec:implications} discusses the implications of these findings for design processes and workflows. Finally, Section~\ref{sec:conclusion} addresses limitations, outlines potential avenues for future work, and concludes with reflections on integrating generative AI into engineering applications.



\begin{figure*}[h!]
    \centering
    \begin{tcolorbox}[colframe=cborange, colback=cborange!5, title=Designer Prompt]
    We are in the conceptual design phase of an estateback car and I want to explore different styling options. I have provided a sketch of the initial design, and I would like to generate high-quality renderings based on specific styling directions. Could you assist in creating variations of the design using the following prompts? 
        
        \begin{minipage}{0.64\textwidth}
        
\begin{itemize}
    \item \textit{A futuristic sports car, aggressive design aerodynamic, vibrant red color.}
    \item \textit{A classic vintage car pearlescent champagne color, chrome details, and a luxurious 1920s style.}
    \item \textit{A rugged off-road SUV, all terrains vehicle, matte black, robust and  muscular design.}
\end{itemize}
        \end{minipage}
        \hspace{1em}
        \begin{minipage}{0.35\textwidth}
            \centering
            \includegraphics[width=0.95\linewidth]{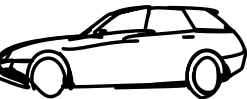}
        \end{minipage}
    \end{tcolorbox}

    \begin{tcolorbox}[colframe=cbgray, colback=cbgray!5, title=AutoGen - Orchestrating AI Design Agents]
Activating the Styling Agent to generate high-quality renderings based on the specified design prompts.
\end{tcolorbox}
    \begin{tcolorbox}[colframe=cbgreen, colback=cbgreen!5, title=Styling Agent Response]
        \begin{minipage}{0.35\textwidth}
        High-resolution renderings have been generated, ensuring accurate representation of the specified aesthetics. 
        \end{minipage}
        \hspace{1em}
        \begin{minipage}{0.64\textwidth}
            \centering
            \includegraphics[width=\linewidth]{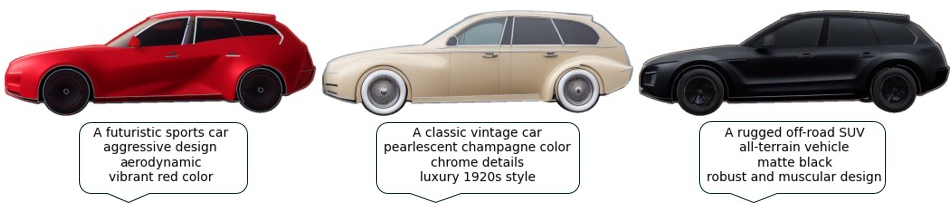}
        \end{minipage}
    \end{tcolorbox}

\begin{tcolorbox}[colframe=cbblue,colback=cbblue!5,title=Engineer Prompt]
    Based on the generated renderings, could you please search the DrivAerNet++ database for similar designs? If you find relevant matches, please also retrieve the corresponding CFD simulation data. This will help in evaluating aerodynamic performance and guiding further refinements. 
\end{tcolorbox}
    \begin{tcolorbox}[colframe=cbgray, colback=cbgray!5, title=AutoGen - Orchestrating AI Design Agents]
Calling the CAD Agent to retrieve similar 3D designs from DrivAerNet++ and the Simulation Agent to provide aerodynamic analysis, including CFD simulation data for performance evaluation.
    \end{tcolorbox}

    \begin{minipage}{0.49\textwidth} 
        \begin{tcolorbox}[colframe=cbgreen, colback=cbgreen!5, title=CAD Agent Response]
            \begin{minipage}{\textwidth}
Retrieved the top 3 designs similar to the estateback sketch from the DrivAerNet++ database with their IDs and STL files. A Blender window will open with the STL files loaded, allowing further inspection and modification of the retrieved designs.
                                \vspace{4pt}

            \end{minipage}
            \hspace{1em}
            \begin{minipage}{\textwidth}
                \centering
                \includegraphics[width=\linewidth]{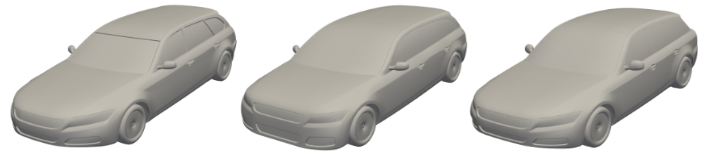}
            \end{minipage}
        \end{tcolorbox}
    \end{minipage}
    \hfill
    \begin{minipage}{0.49\textwidth} 
            \begin{tcolorbox}[colframe=cbgreen, colback=cbgreen!5, title=Simulation Agent Response]
        \begin{minipage}{\textwidth}
Aerodynamic simulation retrieval has been completed for the three designs. Detailed reports have been generated, including visualizations of velocity distribution in the symmetry plane to analyze flow characteristics. A Paraview window will open with the VTK files loaded, allowing further inspection and modification of the retrieved designs.
\vspace{6pt}
\end{minipage}
                \hspace{1em}
        \begin{minipage}{\textwidth}
            \centering
            \includegraphics[width=\linewidth]{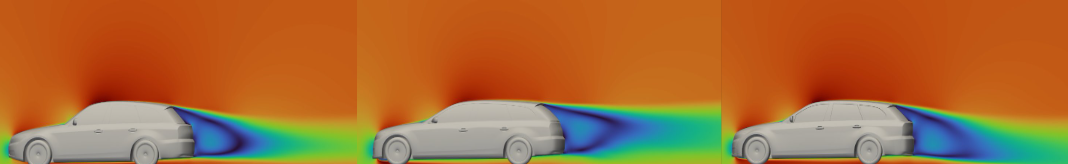}
        \end{minipage}
    \end{tcolorbox}
    
    \end{minipage}

                \vfill





\vspace{5pt}
\caption{Example use case of the AI-driven multi-agent system for car design. Designers start with a 2D sketch, which the Styling Agent uses to generate high-resolution renderings. The CAD Agent then retrieves similar 3D designs from DrivAerNet++ for further exploration. Engineers can request aerodynamic analysis, and the Simulation Agent provides CFD simulation data, allowing for rapid performance evaluation and iterative design improvements. }
\label{fig:DesignAgents_1}
    \end{figure*}

\begin{figure*}[h!]
    \centering
    \begin{tcolorbox}[colframe=cborange, colback=cborange!5, title=Designer Prompt]
I would like to explore intermediate designs between two car configurations. Could you interpolate between the 3D mesh of the estateback and the notchback car, both with smooth underbodies? The goal is to generate a smooth transition between these two designs while preserving aerodynamic integrity.
    \end{tcolorbox}

    \begin{tcolorbox}[colframe=cbgray, colback=cbgray!5, title=AutoGen - Orchestrating AI Design Agents]
    Calling the CAD Agent.
    \end{tcolorbox}
    \begin{tcolorbox}[colframe=cbgreen, colback=cbgreen!5, title=CAD Agent Response]
        \begin{minipage}{0.35\textwidth}
Three intermediate shapes have been generated by interpolating in the latent space. These interpolations smoothly transition between the estateback and notchback designs while maintaining geometric consistency. The generated shapes are now available for further evaluation and aerodynamic analysis.
        \end{minipage}
                        \hspace{1em}
                \begin{minipage}{0.63\textwidth}
            \centering
            \includegraphics[width=\linewidth]{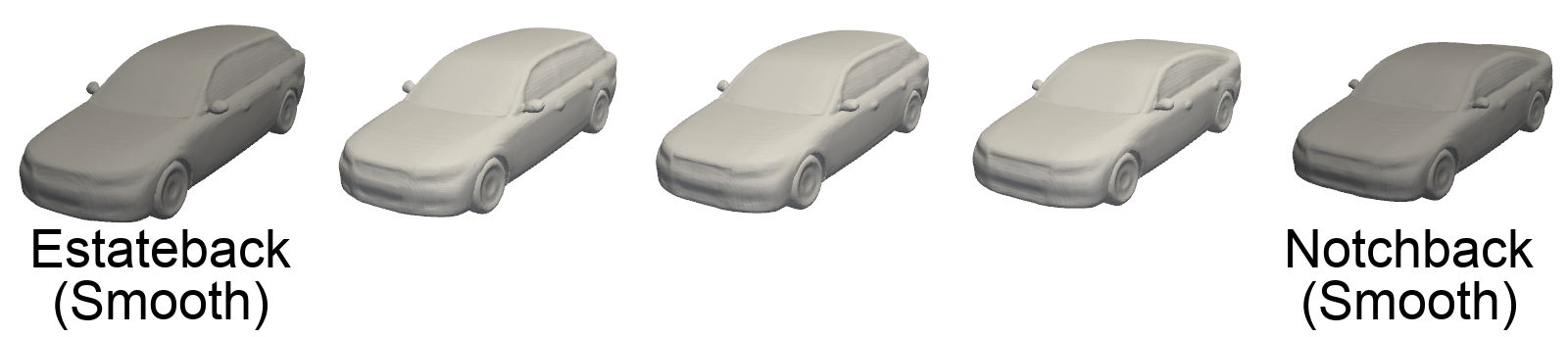}
            \label{fig:enter-label}
        \end{minipage}
\end{tcolorbox}

\begin{tcolorbox}[colframe=cbblue, colback=cbblue!5, title=Engineer Prompt]
Since the interpolated shapes are newly generated and not part of the existing dataset, I would like to run a CFD simulation to analyze their aerodynamic performance. Could you please generate a high-quality computational mesh for the first interpolated design using SnappyHexMesh?
\end{tcolorbox}

\begin{tcolorbox}[colframe=cbgray, colback=cbgray!5, title=AutoGen - Orchestrating AI Design Agents]
Calling the Meshing Agent.
\end{tcolorbox}
         \begin{tcolorbox}[colframe=cbgreen, colback=cbgreen!5, title=Meshing Agent Response]
            \begin{minipage}{\textwidth}
Generated CFD mesh for the specified designs with refinements in critical areas. The computational domain was first initialized using \textit{blockMeshDict}, followed by mesh refinement and surface snapping with \textit{snappyHexMesh}. Finally, \textit{checkMesh} was executed to verify mesh quality, ensuring all key metrics are within acceptable limits. Log files for all steps have been saved.
                \vspace{4pt}
            \end{minipage}
        \end{tcolorbox}

\caption{Continuation of the AI-driven multi-agent system use case for car design. After retrieving similar designs, the CAD Agent generates intermediate shapes through latent space interpolation. These interpolated designs transition smoothly between the estateback and notchback configurations while maintaining geometric consistency. Since these new shapes are not part of the existing dataset, the Meshing Agent generates high-quality computational meshes for aerodynamic analysis. Engineers can then run CFD simulations to evaluate performance, enabling iterative refinement and optimization.}
\label{fig:DesignAgents_2}
\end{figure*}

\section{Related Work}
\label{sec:related_work}

We review existing research and industry trends related to generative AI applications in automotive design. This provides context for understanding how our framework addresses gaps and extends prior work.


The integration of artificial intelligence (AI) into automotive design processes has garnered significant attention, providing innovative solutions for enhancing creativity, performance evaluation, and overall design efficiency. Liu et al. \cite{demirel2024human} introduced a human-centered generative design framework, integrating human factors and mechanical attributes early in the design process. Initial case studies suggested that this multidisciplinary approach yielded more diverse and creative outcomes, demonstrating the potential advantages of synthesizing human-centric considerations within generative design workflows. Similarly, Yuan et al. \cite{yuan2024form} explored AI-driven generative methods for vehicle front-face design, integrating tools such as ChatGPT, Midjourney, and Vega AI. Their methodology emphasized a smooth transition through design phases by systematically blending aesthetic and engineering criteria. Additionally, they applied Kansei engineering and AI-driven form generation to tailor designs specifically to female aesthetics, effectively quantifying subjective preferences and enhancing the workflow with targeted solutions aligned with market demographics.

Ananthan et al.~\cite{ananthan2024machine} introduced a methodology that leveraged machine learning techniques specifically tailored for road vehicle aerodynamics. Their approach enabled designers to rapidly iterate between conceptual designs and aerodynamic performance assessments by optimizing shapes within a learned latent space, effectively connecting early-stage design to aerodynamic outcomes. Similarly, Aréchiga et al.~\cite{arechiga2023drag} proposed drag-guided diffusion models for vehicle image generation, further emphasizing the integration of aesthetics and aerodynamic considerations in automotive design workflows.  Addressing the balance between rapid prototyping and visual fidelity, Radhakrishnan et al. \cite{radhakrishnan2018creative} developed a generative design system that leveraged Generative Adversarial Networks (GANs). Their unique convolutional architecture mimicked designers' sketching style during initial prototyping phases, efficiently producing high-quality, novel car designs from minimal studio sketches, significantly accelerating the conceptual design cycle.

Further enriching design capabilities, Edwards et al. \cite{edwards2024sketch2prototype} presented Sketch2Prototype, a generative AI-based framework that enabled rapid exploration and prototyping in early design phases. This approach highlighted the potential of generative models, specifically in streamlining the iterative process between concept creation and detailed evaluation. In Ref.~\cite{ocker2025idea}, the authors introduced a multi-agent system (MAS) that leveraged vision-language models (VLMs) and specialized agents mimicking distinct engineering roles to collaboratively generate CAD models. Their approach outperformed a single-shot generative baseline, demonstrating enhanced readiness and requirement compliance due to its iterative self-feedback architecture.



However, existing studies primarily face limitations due to the lack of large-scale, multimodal datasets that combine diverse design variations with high-fidelity simulations.  This research gap has constrained the effective coupling of aesthetics and design performance evaluation, limiting the ability to optimize designs holistically. Our work addresses this challenge through the DrivAerNet++ dataset and the introduction of \textit{Design Agents}. By integrating automation, collaboration, and intelligent decision-making, these agents enable rapid design iteration. Traditional car design workflows, from conceptual sketching to simulation, can take several weeks, whereas our multi-agent framework reduces this timeline to minutes by automating 3D shape retrieval, CFD meshing, and aerodynamic evaluation. The \textbf{Design Agents} operate on a large-scale, high-fidelity CFD dataset, distinguishing our approach by utilizing an extensive collection of 8,000 industry-standard car designs and addressing both aerodynamic performance and aesthetic considerations.  Figures~\ref{fig:DesignAgents_1} and~\ref{fig:DesignAgents_2} illustrate a representative use case from our study, demonstrating how engineers and designers can effectively interact with our proposed multi-agent design system.

\begin{figure*}[h!]
    \centering
    \includegraphics[width=\linewidth]{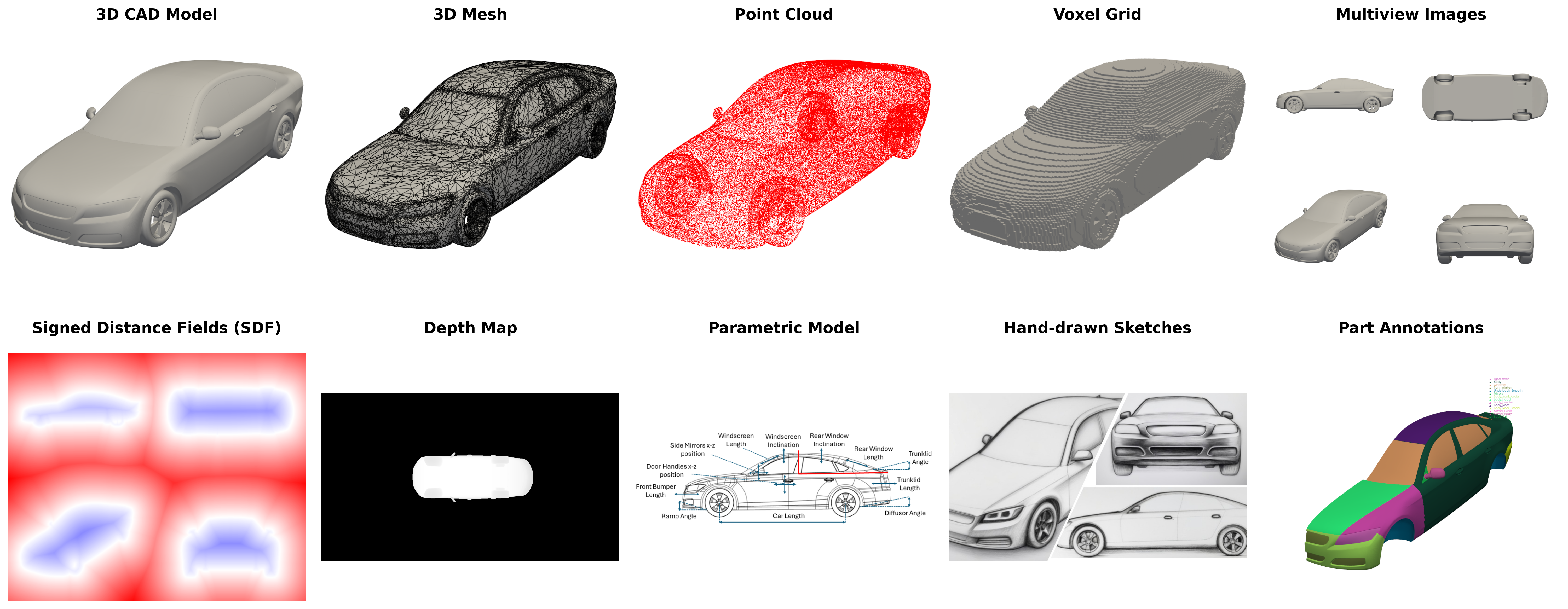}
    \caption{Different data representations and modalities from DrivAerNet++, a dataset comprising 8,000 industry-standard car designs. These modalities—including 3D CAD models, 3D meshes, point clouds, voxel grids, depth maps, and part annotations—are leveraged by various generative AI models depending on the design task, such as retrieval, 3D reconstruction, styling, and aerodynamic simulation. In this work, we further extend DrivAerNet++ by adding Signed Distance Function (SDF) representations, multi-view images, and sketches to support diverse generative design tasks.}
    \label{fig:drivaernet_modalities}
\end{figure*}

\section{Methodology: A Multi-Agent AI Framework for Car Design}
\label{sec:DesignAgents}
This section presents our multi-agent AI framework, explaining how various specialized AI agents collaboratively automate and streamline the entire automotive design process. By seamlessly integrating large language models (LLMs), vision-based latent models such as Stable Diffusion XL~\cite{podell2023sdxl} with ControlNet~\cite{zhang2023adding}, and geometric deep learning models (DeepSDF~\cite{park2019deepsdf}, PointNet~\cite{qi2017pointnet},  RegDGCNN~\cite{wang2019dynamic,elrefaie2024drivaernet,elrefaie2025drivaernet++}, and TripNet~\cite{chen2025tripnet}) using the AutoGen framework~\cite{wu2023autogen}, we demonstrate a collaborative multi-agent system capable of tackling various design tasks.

Our multi-agent system facilitates collaboration between engineers and designers in the car design process. Given a \textit{2D conceptual sketch}, the \textbf{Styling Agent}, powered by SDXL~\cite{podell2023sdxl} and ControlNet~\cite{zhang2023adding}, generates high-resolution renderings with text-driven styling. The \textbf{CAD Agent} retrieves similar 3D designs from DrivAerNet++ or generates new shapes. The \textbf{Meshing Agent} produces CFD-ready meshes and evaluates their quality. Finally, the \textbf{Simulation Agent} predicts aerodynamic properties in real time or retrieves CFD data from DrivAerNet++, enabling rapid design evaluation and optimization.

To orchestrate and coordinate interactions among our proposed AI Design Agents, we leverage AutoGen~\cite{wu2023autogen}, a framework designed for efficient multi-agent collaboration. Within this approach, we specifically examine two essential concepts—\textit{Sequential vs.\ Hierarchical vs.\ Hybrid Agents} and \textit{Cross-Modal Retrieval}—as they critically influence the effectiveness of our agent-based automotive design workflow.


\paragraph{AutoGen: A Framework for Building AI Agents and Applications}

 In our work, we leverage AutoGen~\cite{wu2023autogen}, a framework developed by Microsoft for creating multi-agent AI applications that can act autonomously or work alongside humans. It allows agents to effectively exchange information through natural language prompts, enabling the automation of complex workflows involving diverse software platforms. In our study, Autogen coordinates interactions between specialized design agents, such as CAD retrieval, mesh generation, aerodynamic simulation, and styling agents, streamlining the engineering design process and enhancing productivity and creativity. The integration provided by AutoGen significantly accelerates tasks by reducing manual intervention, facilitating rapid design iterations, and improving collaborative decision-making.

\paragraph{Sequential vs. Hierarchical vs. Hybrid Agents}

Sequential, hierarchical, and hybrid agents represent different structures for AI-driven workflows~\cite{xi2025rise}. Sequential agents operate in a step-by-step manner, where one agent completes a task before passing the output to the next (e.g., a sketching agent generates a 2D concept, which is then refined by a styling agent, followed by a simulation agent for aerodynamic evaluation). Hierarchical agents, on the other hand, follow a structured decision-making approach, where a central agent (or higher-level controller) delegates tasks to specialized sub-agents, ensuring coordination and goal alignment. This approach is beneficial when tasks require complex dependencies and prioritization. Hybrid agents combine both approaches, allowing agents to work both sequentially and in parallel, where certain tasks may run independently while others follow a defined workflow.

In car design, the hybrid agent approach is the most relevant, as the process involves both sequential dependencies (e.g., starting with sketches, then generating 3D shapes, and finally evaluating aerodynamics) and parallel tasks (e.g., styling and aerodynamic simulation can be iterated independently before final validation). This flexibility allows engineers and designers to optimize performance while maintaining creative freedom in shaping car aesthetics.

\paragraph{Cross-Modal Retrieval}
Cross-modal retrieval in DrivAerNet++ enables engineers and designers to access aerodynamic performance data and parametric design information using diverse input modalities. By leveraging deep learning-based retrieval models, a simple 2D hand-drawn sketch of a car can be used to retrieve the most aerodynamically similar 3D designs from the dataset, along with their corresponding high-fidelity CFD simulations. This allows designers to explore optimized shapes that balance aesthetics and performance, while engineers can efficiently search for design variations with similar aerodynamic properties. The integration of image- and text-based queries with engineering data enhances the design workflow, bridging conceptual sketches with performance-driven design evaluation. Figure~\ref{fig:drivaernet_modalities} illustrates the diverse data modalities in DrivAerNet++, including 3D CAD models, meshes, point clouds, voxel grids, depth maps, and part annotations, while also highlighting our extensions with Signed Distance Function (SDF) representations, multi-view images, and sketches to enhance generative design tasks.


\subsection{Enhancing Aesthetic Exploration with the Styling Agent}

In this section, we introduce the \textit{Styling Agent} to demonstrate an agentic way of providing design inspirations in automotive design workflows. Given a hand-drawn sketch accompanied by a text prompt, our Styling Agent leverages advanced generative AI models to produce high-resolution, photorealistic renderings of car designs. This enables designers to rapidly visualize diverse aesthetic concepts that are inspired by their sketches and iterate creatively in early design phases.

To demonstrate our Styling Agent, we required a dataset of car sketches. As no public dataset exists, we created our own dataset based on the DrivAerNet++ dataset. Curating a dataset of 8,000 hand-drawn sketches that correspond to cars in the DrivAerNet++ dataset is challenging due to the labor-intensive nature of manual sketching and the variability in artistic styles. Therefore, we employ two automated approaches to overcome this challenge: a traditional computer vision method using Canny edge detection~\cite{canny1986computational} and a pre-trained generative AI approach leveraging CLIPasso model~\cite{vinker2022clipasso} to produce sketch abstractions from input images. Prior work has shown that the output of AI generated sketches aligns well with human-drawn sketches. Unlike methods requiring explicit sketch datasets for training, our approach generates sketches\footnote{CLIPasso and Canny edge detection are used solely for generating a dataset of hand-drawn sketches to train and evaluate shape and data retrieval models. The agent itself does not utilize these methods during operation. Instead, it performs retrieval based on user-provided sketches, under the assumption that human-drawn sketches resemble those generated by CLIPasso or Canny edge detection.} that closely resemble those drawn by designers in early-stage conceptual design, ensuring consistency and scalability in capturing design intent.

\begin{figure}[h!]
    \centering
    \begin{subfigure}{0.3\linewidth}
        \centering
        \includegraphics[width=\textwidth]{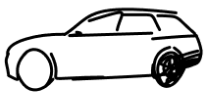}
        \caption{Generated with 1 sketch, 25 strokes.}
        \label{fig:1_25}
    \end{subfigure}
    \begin{subfigure}{0.3\linewidth}
        \centering
        \includegraphics[width=\textwidth]{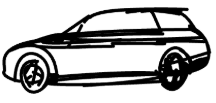}
        \caption{Generated with 1 sketch, 50 strokes.}
        \label{fig:1_50}
    \end{subfigure}
    \begin{subfigure}{0.3\linewidth}
        \centering
        \includegraphics[width=\textwidth]{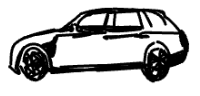}
        \caption{Generated with 1 sketch, 100 strokes.}
        \label{fig:1_100}
    \end{subfigure}
    \hfill
    \begin{subfigure}{0.3\linewidth}
        \centering
        \includegraphics[width=\textwidth]{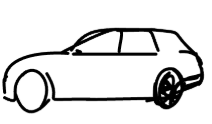}
        \caption{Generated with 3 sketches, 25 strokes.}
        \label{fig:3_25}
    \end{subfigure}
    \begin{subfigure}{0.3\linewidth}
        \centering
        \includegraphics[width=\textwidth]{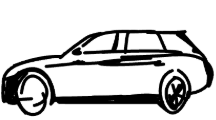}
        \caption{Generated with 3 sketches, 50 strokes.}
        \label{fig:3_50}
    \end{subfigure}
    \begin{subfigure}{0.3\linewidth}
        \centering
        \includegraphics[width=\textwidth]{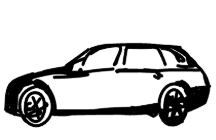}
        \caption{Generated with 3 sketches, 100 strokes.}
        \label{fig:3_100}
    \end{subfigure}
    \vspace{2pt}
    \caption{Sketches of an Estateback car from the DrivAerNet++ dataset~\cite{elrefaie2025drivaernet++}, generated using CLIPasso~\cite{vinker2022clipasso}. The sketches vary based on the number of strokes and input sketches used, demonstrating the model's ability to capture design features with different levels of abstraction.}
    \label{fig:clipasso_estateback}
\end{figure}

\paragraph{Sketch Generation Using CLIPasso}
This subsection shows how we leverage the CLIPasso model~\cite{vinker2022clipasso} to generate expressive and representative sketch abstractions from realistic car images, facilitating early-stage design exploration.
The CLIPasso model is a framework for generating sketches using differentiable vector graphics guided by neural optimization~\cite{vinker2022clipasso}. It leverages CLIP (Contrastive Language-Image Pre-training) \cite{radford2021learning}, a vision-language model, to iteratively refine sketches based on textual descriptions or visual concepts. By optimizing Bézier curves directly in a differentiable manner, CLIPasso efficiently synthesizes minimalistic yet expressive sketches that closely match the desired semantic content~\cite{vinker2022clipasso}. This approach bridges neural rendering techniques and vector graphics, providing a precise and intuitive method for sketch-based conceptual design tasks and interactive creativity support~\cite{vinker2022clipasso,frans2021clipdraw}.

Figure~\ref{fig:clipasso_estateback} presents sketches of an Estateback car from the DrivAerNet++ dataset, showcasing variations based on the number of input sketches and the number of strokes used for abstraction. The top row depicts sketches generated with a single sketch, while the bottom row shows those generated using three sketches, with each column corresponding to an increasing number of strokes. Here, increasing the number of strokes enhances the quality and level of detail in the sketch, allowing for a more precise representation of design features.
\begin{figure*}[h!]
    \centering
    \includegraphics[width=\linewidth]{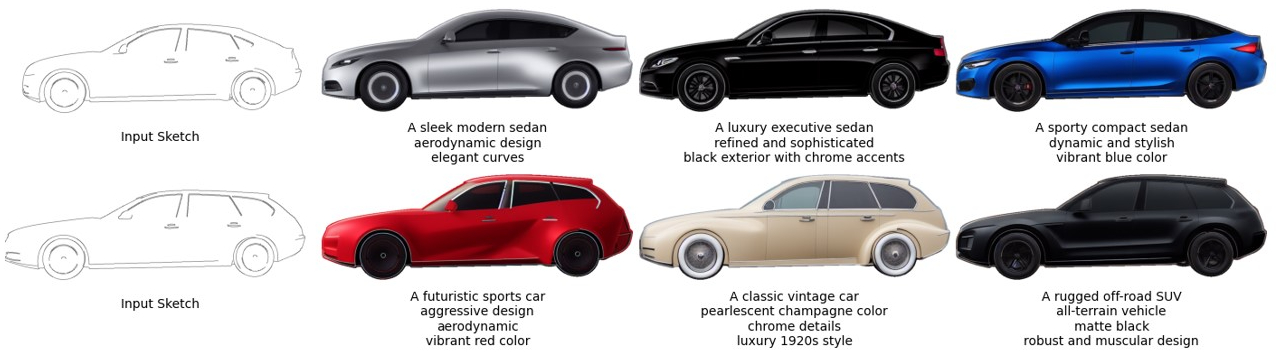}
    \caption{Input and output of the \textit{Styling Agent}. The input consists of a hand-drawn sketch and a text prompt, which guides the model to generate diverse car renderings. The results demonstrate variations in styling, materials, and lighting based on different text prompts. Two car categories are shown: sedan (fastback) in the top row and SUV (estateback) in the bottom row, highlighting how AI-driven design can adapt to user preferences and generate high-quality, photorealistic concepts.}
    \label{fig:car_styling_sdxl}
\end{figure*}
\paragraph{ControlNet for Guided Conceptual Design Generation}
ControlNet~\cite{zhang2023adding} is a conditional image generation architecture built upon diffusion-based generative models. It incorporates external guidance signals—such as edge detection maps, semantic segmentations, or user-generated sketches—to precisely control the generative process. By embedding these explicit control signals into diffusion models, ControlNet enables designers to transform preliminary sketches into refined, high-resolution renderings while preserving original stylistic and structural intent. In the context of automotive design, ControlNet's controlled generation process facilitates accurate and rapid conceptual exploration, significantly streamlining iterative design refinement and enhancing alignment between stylistic concepts and engineering requirements.

\paragraph{High-resolution and Photorealistic Renderings of Car Designs} 
This subsection introduces the process of using vision-language models, such as Stable Diffusion XL (SDXL)~\cite{podell2023sdxl} and ControlNet~\cite{zhang2023adding}, to produce photorealistic and stylistically diverse car renderings conditioned on user sketches and text prompts. After generating the sketches using either Canny edge detection~\cite{canny1986computational} and CLIPasso~\cite{vinker2022clipasso} for abstraction-based sketch generation, these sketches serve as conditioning inputs for a ControlNet model~\cite{zhang2023adding}, which refines text-driven generation using Stable Diffusion v1.5~\cite{rombach2022high,podell2023sdxl}. We use a combination of base car design prompts, color variations, and stylistic attributes to introduce diverse and creative designs. The generated images are then post-processed to remove the background using morphological operations and contour-based segmentation. The final outputs, along with metadata including the applied prompts, are stored for further analysis. This approach can allow designers to explore multiple aesthetic variations while maintaining the underlying structural integrity of the design. To introduce stylistic diversity, we experiment with different car categories and artistic influences by modifying design prompts. 
Table \ref{tab:style_variations} summarizes the various styling and color variations applied to generate a diverse dataset of rendered car designs. By leveraging AI-driven design agents, we synthesized 8,000 rendered car designs, each incorporating different stylistic influences and color schemes to enhance visual diversity. This approach allows for a broad exploration of aesthetic variations, ensuring that the generated dataset captures a wide spectrum of design possibilities across multiple car categories.

\begin{table}[h!]
    \footnotesize
    \centering
    \begin{tabular}{c@{\hspace{0.1cm}}c@{\hspace{0.1cm}}c} 
        \hline
        \textbf{Category} & \textbf{Style Applied} & \textbf{Color Variation} \\
        \hline
        Futuristic Sports Car & Cyberpunk, High-Tech & Metallic Silver, Neon Blue \\
        Classic Vintage Car & Steampunk, Luxury & Deep Red, Matte Gold \\
        Off-Road Vehicle & Rugged, Urban & Earthy Brown, Matte Gray \\
        Electric Concept Car & Minimalist, Futuristic & Pearl White, Electric Green \\
        Race Car & Aerodynamic, Sporty & Glossy Black, Sunset Orange \\
        City Car & Urban, Eco-Friendly & Midnight Purple, Bright Yellow \\
        Retro-Inspired Car & Vintage, Artistic & Vibrant Blue, Pastel Pink \\
        \hline
    \end{tabular}
        \vspace{2pt}

    \caption{Style and color variations used for enhanced diversity.}
    \label{tab:style_variations}
\end{table}

 We show in Figure \ref{fig:car_styling_sdxl} the transformation of hand-drawn sketches into high-quality car renderings using the \textit{Styling Agent}. The input consists of a 2D sketch obtained from either Canny or CLIPasso, combined with text prompts that guide the generation process. The results showcase diverse styling variations, material finishes, and lighting effects, adapting to different design intents. Two car categories are depicted: fastback sedans (top row) and estateback SUVs (bottom row), illustrating how AI-driven design enables rapid exploration of aesthetic concepts while maintaining structural consistency.

\subsection{Bridging 2D Sketches and 3D Geometries with the CAD Agent}

In this section, we introduce the \textit{CAD Agent}\footnote{To avoid confusion, we differentiate it from the Meshing Agent, which focuses exclusively on generating computational meshes for CFD simulations.}, which bridges conceptual sketches and detailed 3D car designs. Given a 2D hand-drawn sketch, the CAD Agent searches the DrivAerNet++ database in real-time to retrieve the most similar existing 3D design. Beyond retrieval, it can perform generative tasks, synthesizing novel 3D designs and interpolating between two input models to explore intermediate shapes, all conditioned on industry-standard automotive geometries. This facilitates rapid exploration and generation of feasible, production-oriented concepts.


\begin{figure*}[h!]
    \centering
    \includegraphics[width=\linewidth]{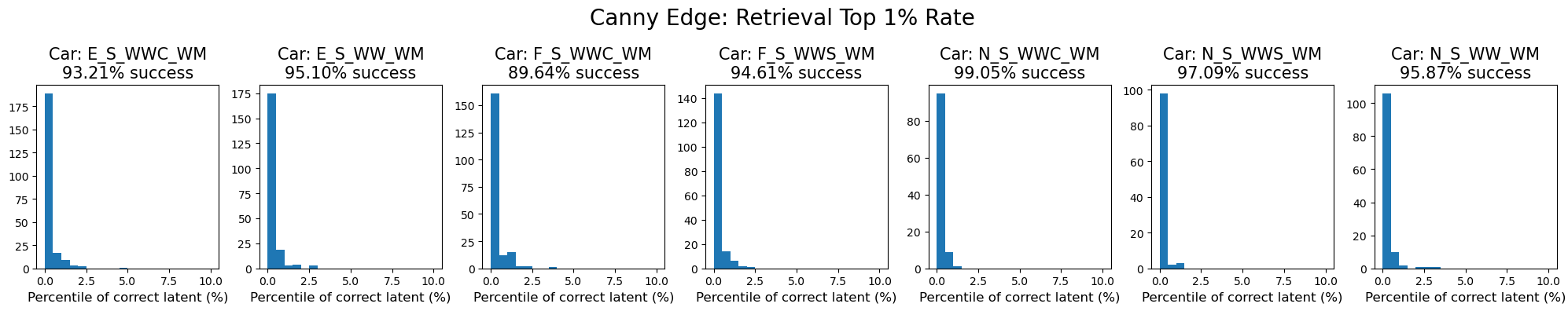}
    \includegraphics[width=\linewidth]{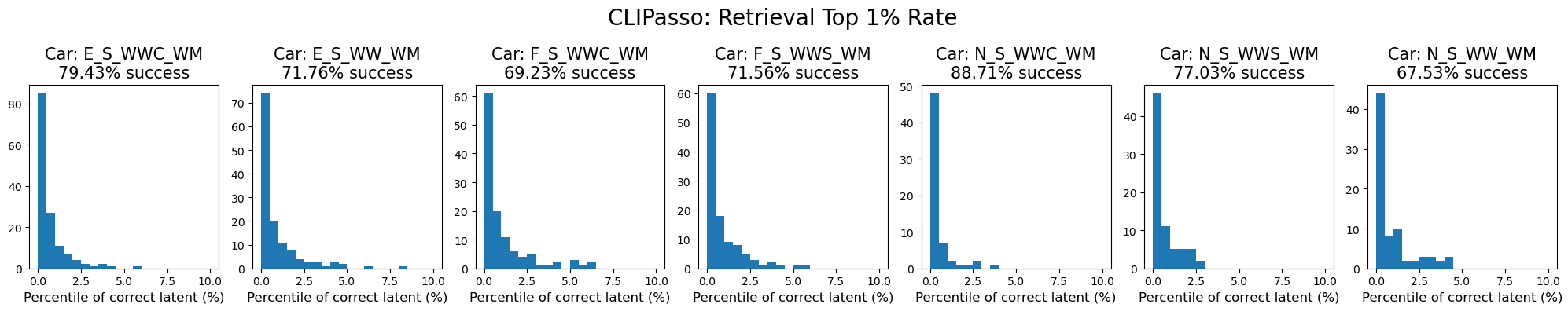}
    \vspace{1pt}
    \caption{Retrieval success rate of our CAD agent for different car categories in the DrivAerNet++ dataset. The $x$-axis represents the percentile rank of the correctly retrieved latent code within the search space, while the $y$-axis shows the frequency of retrieval attempts. Each subplot corresponds to a different car configuration, with the success rate indicating the percentage of cases where the correct latent code was ranked among the top 1\% of retrieved candidates. The results demonstrate the effectiveness of our retrieval model in accurately identifying the closest matching 3D car mesh from a given input sketch.}
    \label{fig:deepsdf_retrieving}
\end{figure*}

\subsubsection{Deep Implicit Representations for Automotive Shapes}
Below we outline our approach using DeepSDF to encode automotive shapes into continuous implicit representations, enabling efficient shape retrieval and generative modeling. DeepSDF~\cite{park2019deepsdf} is a neural implicit representation technique that encodes 3D shapes into a continuous Signed Distance Function (SDF) space, allowing for efficient shape reconstruction and interpolation. Our approach extends DeepSDF for car design by encoding car meshes into a latent space representation while introducing modifications to improve training efficiency and reconstruction quality. Recent research has leveraged DeepSDF for 3D car generation and customization. Morita et al.~\cite{morita2024vehiclesdf} introduced VehicleSDF, enabling the generation of diverse 3D car models while estimating aerodynamic performance via surrogate modeling. Similarly, Miao et al.~\cite{miao2024manipulating} proposed a framework for precise, attribute-specific modifications to both stylistic and geometric aspects of car designs, ensuring structural integrity.

\paragraph{DeepSDF Training Data, Network Architecture and Modifications}

To train our CAD Agent, we require a dataset that includes SDF samples which are essential for training implicit neural shape representations. To generate the training data, we normalize each car mesh to fit within a unit sphere and sample 500,000 spatial points, with a higher concentration near the surface. Each point $(x, y, z)$ is associated with an SDF value that represents the signed distance to the nearest surface.

DeepSDF uses an auto-decoder framework, where a latent vector $\mathbf{z} \in \mathbb{R}^{d}$ is learned for each training shape. The network consists of a multi-layer perceptron (MLP) with eight fully connected layers of 512 neurons each, utilizing ReLU activations. Given an input coordinate $\mathbf{x} \in \mathbb{R}^{3}$ and the latent vector representation $\mathbf{z} \in \mathbb{R}^{d}$, the model predicts the corresponding SDF value $f(\mathbf{x}, \mathbf{z})$.

To improve training efficiency, accuracy, and performance specific to car designs, we introduce the following modifications:
\begin{itemize}
    \item \textbf{Positional Encoding:} Inspired by recent advancements in neural implicit representations~\cite{miao2024manipulating}, we apply a Fourier feature mapping to input coordinates:
    \begin{equation}
        \gamma(\mathbf{x}) = \left( \sin(2^k \pi \mathbf{x}), \cos(2^k \pi \mathbf{x}) \right), \quad k=0, \dots, 8
    \end{equation}
    This transformation aids in learning high-frequency geometric details.
    \item \textbf{Reduced Latent Space Dimension:} The latent vector dimension is reduced from 256 to 16 to optimize storage and improve generalization in the constrained design space of automotive shapes.
\end{itemize}

\paragraph{Training Procedure and Reconstruction}

The model is trained on SDF samples using L1 loss on predicted SDF values, with L2 regularization applied to the latent codes. During training, both the model parameters and latent codes are optimized simultaneously using the Adam optimizer. The loss function is defined as

\begin{equation}
    \mathcal{L}(f_{\theta}(\mathbf{x}), s) = \left| \text{clamp}(f_{\theta}(\mathbf{x}), \pm\delta) - \text{clamp}(s, \pm\delta) \right|
\end{equation}

where $\text{clamp}(x, \delta) := \min(\delta, \max(-\delta, x))$ ensures that the model focuses its representational ability near the surface. 
The total loss for optimizing the model and latent codes is:

\begin{equation}
    \arg \min_{\theta, \{z_i\}_{i=1}^{N}} \sum_{i=1}^{N} \left( \sum_{j=1}^{K} \mathcal{L}(f_{\theta}(z_i, \gamma(\mathbf{x}_j)), s_j) + \lambda \|z_i\|_2^2 \right)
\end{equation}

Given a target shape, we find the associated latent code by using the Adam optimizer to minimize the reconstruction loss. Then, we query the DeepSDF model at spatial points $(x, y, z)$ and extract the final mesh using the Marching Cubes algorithm~\cite{lorensen1998marching}.
To explore the learned latent space, we perform interpolation between latent representations of different car designs, enabling smooth transitions between shape variations.

\subsubsection{Retrieval of 3D Car Models from 2D Sketches}
Here we demonstrate how the CAD Agent retrieves the most similar 3D car geometries from DrivAerNet++ based on input sketches.

\paragraph{Latent-Code-Based Shape Retrieval}
This subsection describes how latent space representations (from DeepSDF) enable efficient identification of matching car geometries based on sketch inputs.
Our approach involves training a Convolutional Neural Network (CNN) to predict the latent code associated with each car mesh based on the input sketch. The CNN takes a sketch as input and outputs the predicted DeepSDF latent code. The model is trained to minimize the L2 loss between the predicted latent code and the ground truth latent code.
We use a single CNN model for all car classes. During retrieval, the input consists of a car class and a sketch, and the output is the car mesh with the closest matching latent code, determined by minimizing the L2 distance between the predicted latent code and the latent codes of meshes within the specified car class:

\begin{equation}
    \hat{z} = f_{\theta}(\text{sketch}, c)
\end{equation}

\begin{equation}
    \text{mesh} = \arg\min_{\text{mesh} \in \mathcal{M}_c} \left\| z_{\text{mesh}} - \hat{z} \right\|_2
\end{equation}

where:
\begin{itemize}
    \item \( f_{\theta} \) is a CNN trained to predict the DeepSDF latent code from a given sketch.
    \item \( c \) is the specified car class.
    \item \( \hat{z} \) is the predicted latent code of the input sketch.
    \item \( z_{\text{mesh}} \) represents the latent codes of available 3D car meshes.
    \item \( \mathcal{M}_c \) is the set of all meshes belonging to the car class \( c \).
    \item The retrieval process selects the mesh whose DeepSDF latent code minimizes the L2 distance to the predicted latent code \( \hat{z} \).
\end{itemize}

The retrieval success rates for different car configurations within the DrivAerNet++ dataset are depicted in Figure~\ref{fig:deepsdf_retrieving}. The $x$-axis represents the percentile rank of the correctly retrieved latent code within the search space, while the $y$-axis indicates the frequency of retrieval attempts. Each subplot corresponds to a specific car configuration, demonstrating the effectiveness of our retrieval model in accurately identifying the closest matching 3D car mesh from a given input.

\begin{figure*}[h!]
    \centering
    \begin{minipage}{0.49\linewidth}
        \centering
        \includegraphics[width=\linewidth]{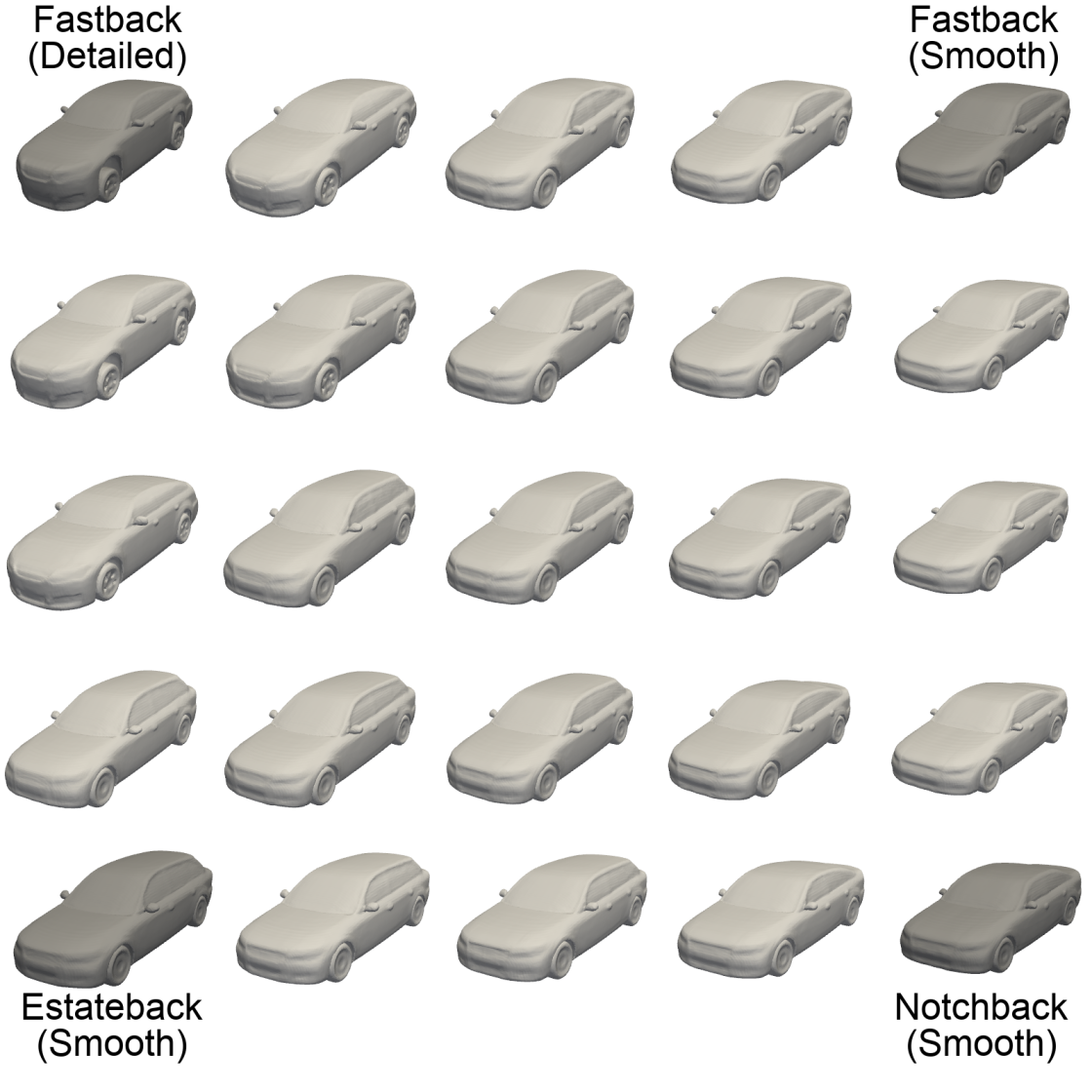}
 
    \end{minipage}
    \hfill
    \begin{minipage}{0.49\linewidth}
        \centering
        \includegraphics[width=\linewidth]{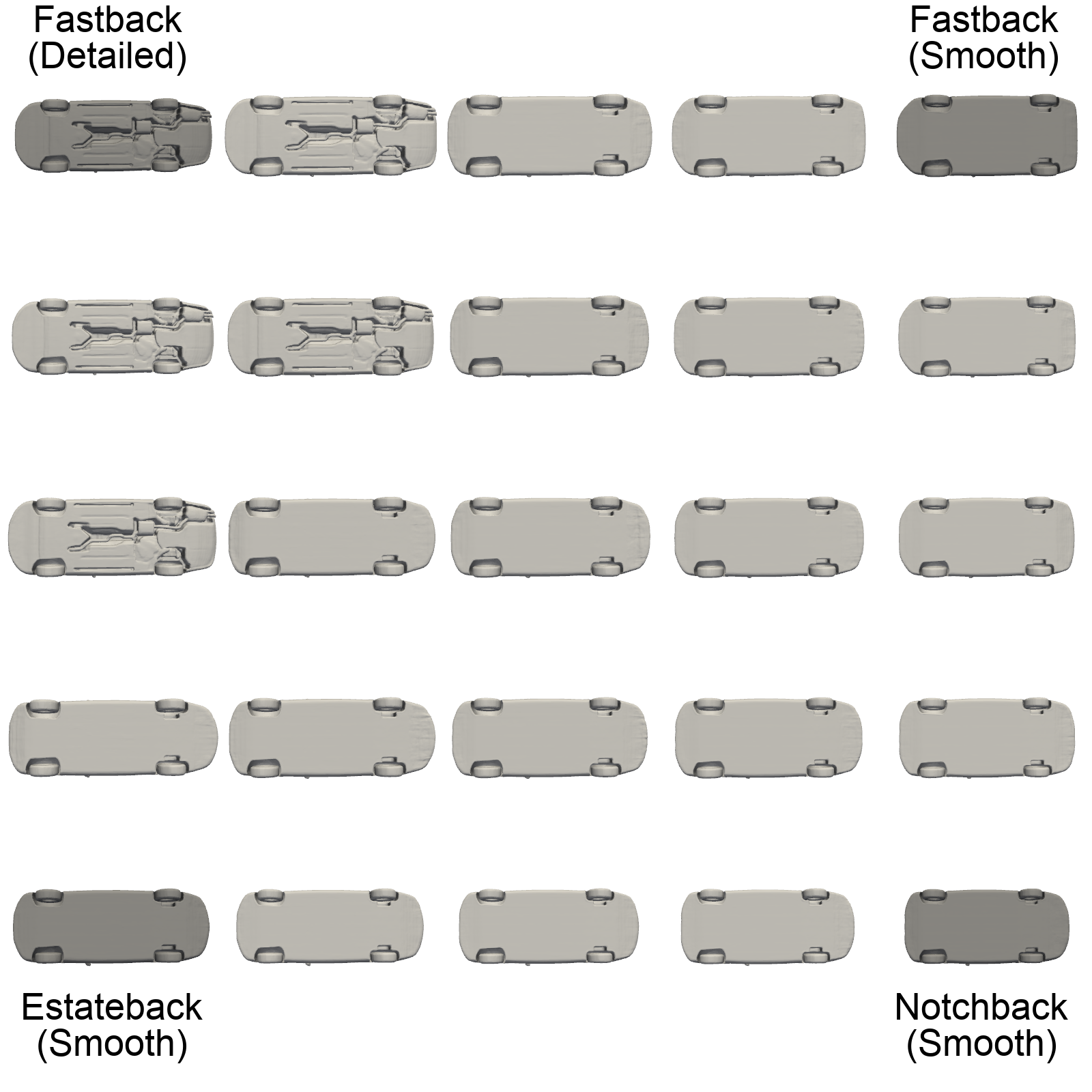}
    \end{minipage}
    \vspace{5pt}
           \caption{Interpolation between four different car designs using our modified DeepSDF. The original input meshes (dark grey) include fastback with detailed underbody, fastback with smooth underbody, estateback, and notchback configurations, while the interpolated meshes (light grey) are generated by interpolating in DeepSDF's latent space. The grid showcases smooth transitions between these shape representations, demonstrating DeepSDF’s capability to generate intermediate geometries by blending latent representations. This structured exploration of the design space enables shape optimization and generative design for aerodynamic and stylistic variations in automotive applications.}
           \label{fig:deep_sdf_interpolation}

\end{figure*}

\paragraph{Feature-Based Retrieval via Cosine Similarity}
Below, we discuss our alternative approach for shape retrieval, using cosine similarity metrics on CNN-extracted image features to robustly identify similar car designs. The similarity between a query image and a set of database images is computed using \textit{cosine similarity} in a high-dimensional feature space derived from a pre-trained ResNet-50~\cite{he2016deep} convolutional neural network. The final fully connected classification layer is removed, yielding a feature extraction model $\mathcal{F}: \mathbb{R}^{H \times W \times C} \rightarrow \mathbb{R}^{n}$, where $H$, $W$, and $C$ represent the height, width, and number of channels of the input image, respectively, and $n$ is the dimensionality of the resulting feature vector. Let the query image be denoted as $\mathbf{I}_q$ and its corresponding feature representation as $\mathbf{q} = \mathcal{F}(\mathbf{I}_q)$. Similarly, let the $i^{th}$ image in the database, $\mathbf{I}_i$, be represented by its feature vector $\mathbf{d}_i = \mathcal{F}(\mathbf{I}_i)$.

The cosine similarity score $S(\mathbf{q}, \mathbf{d}_i)$ is computed as follows:

\[
S(\mathbf{q}, \mathbf{d}_i) = \frac{\sum_{j=1}^{n} q_j d_{i,j}}{\sqrt{\sum_{j=1}^{n} q_j^2} \sqrt{\sum_{j=1}^{n} d_{i,j}^2}}
\]

where $q_j$ and $d_{i,j}$ represent the $j^{th}$ component of the query and database feature vectors, respectively. This similarity measure captures the cosine of the angle between the two vectors in the $n$-dimensional feature space, ensuring invariance to vector magnitudes. The similarity scores are aggregated into a vector $\mathbf{S} \in \mathbb{R}^{M}$, where $M$ denotes the number of images in the database.

To ensure a meaningful retrieval process, we apply two sequential filtering criteria: (1) rank the similarity scores in descending order such that $S(\mathbf{q}, \mathbf{d}_{i_1}) \geq S(\mathbf{q}, \mathbf{d}_{i_2}) \geq \cdots \geq S(\mathbf{q}, \mathbf{d}_{i_M})$, and (2) validate the top-$K$ ranked database images to ensure the existence of corresponding 3D geometries (STL files). Mathematically, this can be expressed as:

\[
\mathcal{R}_K = \{ \mathbf{d}_{i_k} \,|\, k \in [1, K], \, \exists \, \mathcal{G}_{i_k} \in \mathcal{S} \}
\]

where $\mathcal{R}_K$ represents the final set of $K$ retrieved images, $\mathcal{G}_{i_k}$ is the 3D STL geometry associated with the $k^{th}$ ranked image, and $\mathcal{S}$ denotes the set of all valid STL files.

The resulting set $\mathcal{R}_K$ consists of the top-$K$ most similar database images, validated for 3D geometry consistency. Each retrieval result is accompanied by its corresponding similarity score and STL file, enabling subsequent visualization and downstream tasks such as shape analysis and geometric inspection.

\subsubsection{Generating and Exploring New Designs in 3D Latent Spaces}
Here, we illustrate the generative capabilities of the CAD Agent in synthesizing novel, intermediate automotive shapes by smoothly interpolating in learned latent spaces of DeepSDF.
Given the latent codes of two or more car models, a weighted average of these codes produces a new latent representation that blends features of the input designs. When decoded by the DeepSDF network, the resulting shapes are visually plausible and maintain realism, showcasing the latent space's ability to capture meaningful geometric relationships and support novel design generation.

Figure~\ref{fig:deep_sdf_interpolation} demonstrates the effectiveness of our modified DeepSDF in generating smooth transitions between various car designs, such as estateback, fastback with smooth underbody, fastback with detailed underbody, and notchback configurations. This approach allows for conditioning on two designs and interpolating intermediate designs, facilitating a structured exploration of the design space for aerodynamic and stylistic variations in automotive applications. 


Our CAD Agent leverages a learned latent space to facilitate efficient 3D car shape generation, retrieval, and interpolation, enabling smooth, high-quality reconstructions essential for automotive design workflows. By utilizing these implicit shape representations, our framework effectively generates diverse geometries and seamlessly interpolates between existing designs, significantly enhancing the flexibility and creativity of AI-assisted automotive design tasks.

\begin{figure*}[h!]
    \centering
    \begin{tcolorbox}[colframe=cbblue,colback=cbblue!5,title=Engineer Prompt]
        Generate a high-quality computational mesh for an estateback car in OpenFOAM. Set up a virtual wind tunnel with blockMeshDict, split the domain across 128 CPU-cores, and run SnappyHexMesh for refinement. Then, execute checkMesh to verify quality and plot the mesh in the y-symmetry plane. 
    \end{tcolorbox}

    \begin{tcolorbox}[colframe=cbgreen,colback=cbgreen!5,title=Meshing Agent Response]
        \begin{minipage}{0.6\textwidth}
            The virtual wind tunnel has been set up using blockMeshDict, and the computational domain has been parallelized over 128 CPU cores. SnappyHexMesh was executed with castellated meshing, snapping, and boundary layer addition to refining the mesh around the car body, wheels, and underbody. A refinement region was applied to improve wake resolution, and checkMesh verified that all quality metrics are within acceptable limits. The final mesh has been visualized in the y-symmetry plane and is ready for CFD simulations. Log files are saved. 
        \end{minipage}
        \hspace{1em}
        \begin{minipage}{0.35\textwidth}
            \centering
            \includegraphics[width=\linewidth]{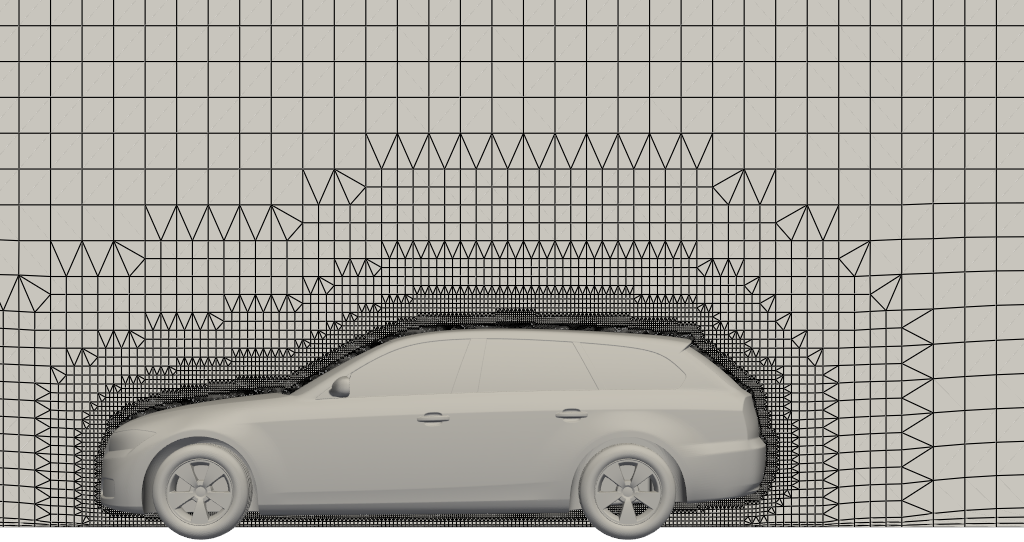}
        \end{minipage}
    \end{tcolorbox}

    \vspace{0.5em}

    \begin{tcolorbox}[colframe=cbblue,colback=cbblue!5,title=Engineer Prompt]
        The transition between layers around the car body and away from the car is kinda coarse. Could you please add an extra refinement layer or a volume refinement region to reduce large variations in cell sizes?
    \end{tcolorbox}

    \begin{tcolorbox}[colframe=cbgreen,colback=cbgreen!5,title=Meshing Agent Response]
        \begin{minipage}{0.6\textwidth}
        An additional volume refinement region has been introduced to minimize abrupt cell size variations. The total cell count increased by 17.11\%, maintaining a well-structured mesh without excessive clustering. The maximum aspect ratio and non-orthogonality remain within limits, ensuring stability for CFD simulations. The mesh quality has been checked, and all metrics remain within acceptable limits. 
        \end{minipage}
        \hspace{1em}
        \begin{minipage}{0.35\textwidth}
            \centering
            \includegraphics[width=\linewidth]{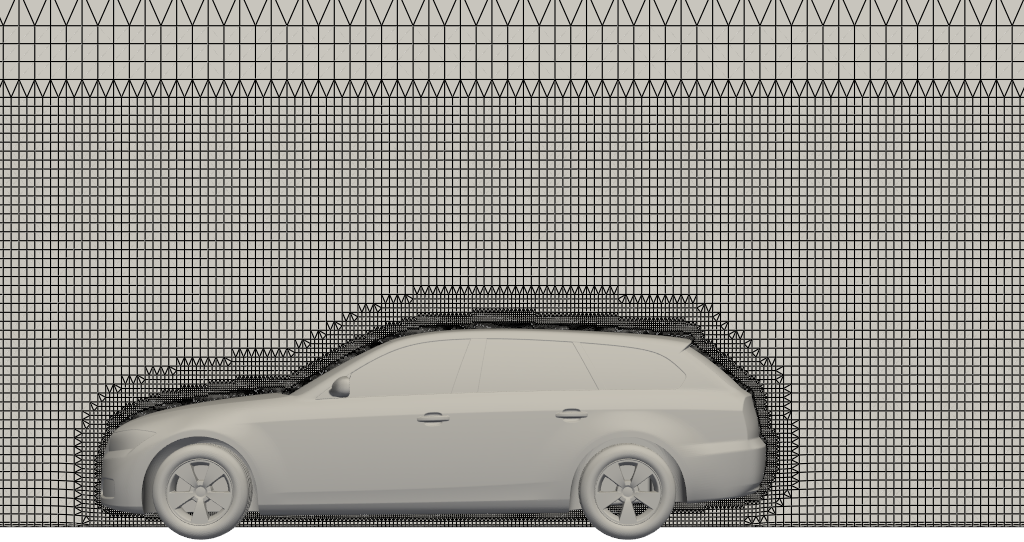}
        \end{minipage}
    \end{tcolorbox}

    \vspace{0.5em}

    \begin{tcolorbox}[colframe=cbblue,colback=cbblue!5,title=Engineer Prompt]
        The wake region does not appear refined enough. I'm concerned that the wake flow and flow field separation might not be captured correctly. Can we improve this?
    \end{tcolorbox}

    \begin{tcolorbox}[colframe=cbgreen,colback=cbgreen!5,title=Meshing Agent Response]
        \begin{minipage}{0.6\textwidth}
Two additional volumes of refinement have been added in the wake region to improve resolution and better capture flow separation. As a result, the total number of cells in the third mesh increased by approximately $237.75\%$ compared to the first mesh and $188.40\%$ compared to the second mesh.

        \end{minipage}
        \hspace{1em}
        \begin{minipage}{0.35\textwidth}
            \centering
            \includegraphics[width=\linewidth]{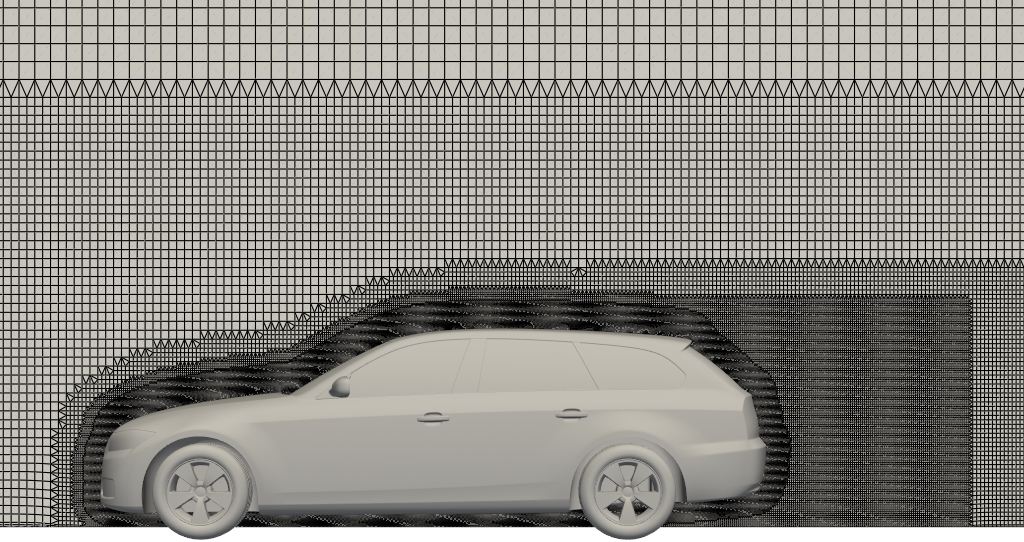}
        \end{minipage}
    \end{tcolorbox}

\caption{Example conversation between an engineer and the Meshing Agent, refining the computational mesh for improved aerodynamic accuracy. The Meshing Agent can run OpenFOAM commands, post-process results, and analyze mesh quality to ensure optimal refinement for accurate simulations.}
    \label{fig:ai-prompts-sketches}
\end{figure*}

\subsection{Automating Computational Meshing with the Meshing Agent}
In this section, we introduce the \textit{Meshing Agent}, responsible for generating high-quality computational meshes from 3D car designs to enable accurate CFD simulations. Given a 3D car model and text prompts specifying meshing requirements, the Meshing Agent interacts with OpenFOAM's \textit{snappyHexMesh} utility, automatically producing refined computational meshes. The agent further verifies mesh quality through standard checks and iteratively improves mesh fidelity, ensuring optimal performance for subsequent aerodynamic analyses.

Recent advancements in leveraging LLMs have demonstrated promising results in automating and enhancing CFD simulations. For instance, Pandey et al.~\cite{pandey2025openfoamgpt} introduced OpenFOAMGPT, a retrieval-augmented generation (RAG) agent tailored for OpenFOAM-based CFD tasks, showcasing its ability to handle complex simulations efficiently. Similarly, Xu et al.~\cite{xu2024llm} developed an LLM agent for fire dynamics simulations, enhancing the usability of FireFOAM by enabling natural language interactions for case configuration and simulation evaluation. Additionally, Chen et al.~\cite{chen2024metaopenfoam} proposed MetaOpenFOAM, an LLM-based multi-agent framework designed to automate CFD simulations using natural language inputs, further streamlining the simulation workflow.

SnappyHexMesh~\cite{openfoam2023snappyhexmesh} is an advanced mesh-generation utility in OpenFOAM~\cite{OpenFOAMv11}, a C++-based computational fluid dynamics software, specifically designed to generate computational meshes from complex geometries, widely utilized in automotive aerodynamic simulations. It employs a robust hex-dominant meshing technique, capable of accurately capturing intricate geometry details by refining mesh cells around surfaces and within specified regions~\cite{openfoam2023snappyhexmesh}. This method allows precise representation of complex automotive geometries, enabling high-fidelity simulations. Mesh generation typically constitutes the most time-consuming step in CFD workflows and significantly impacts the quality and accuracy of simulation results. Thus, efficient and accurate meshing is essential for reliable aerodynamic analysis.

\paragraph{Interactive Mesh Generation and Refinement} 
For the Meshing Agent, we utilize OpenAI's API (version 0.28) with the GPT-3.5-turbo model~\cite{openai2023gpt35turbo} to integrate LLMs with Python scripts, automating CFD simulations using OpenFOAM. This setup enables the interaction between natural language queries and engineering workflows, allowing LLMs to execute OpenFOAM commands, configure solver settings, and retrieve simulation results. Additionally, SnappyHexMesh is employed for generating high-quality CFD meshes, ensuring accurate aerodynamic analysis while streamlining the meshing and simulation pipeline. Figure~\ref{fig:ai-prompts-sketches} illustrates an example conversation between an engineer and the Meshing Agent, refining the computational mesh for improved aerodynamic accuracy. The Meshing Agent is capable of executing various OpenFOAM utilities, such as generating the computational domain with \textit{blockMeshDict}, partitioning the mesh using \textit{decomposeParDict} for parallel execution, modifying and running \textit{snappyHexMesh} for mesh refinement, interpreting the output of \textit{checkMesh}, and visualizing the mesh quality and refinement regions. This integrated capability ensures efficient and robust preparation of computational meshes, crucial for accurate CFD simulations.




\begin{figure*}[h!]
    \includegraphics[width=\linewidth]{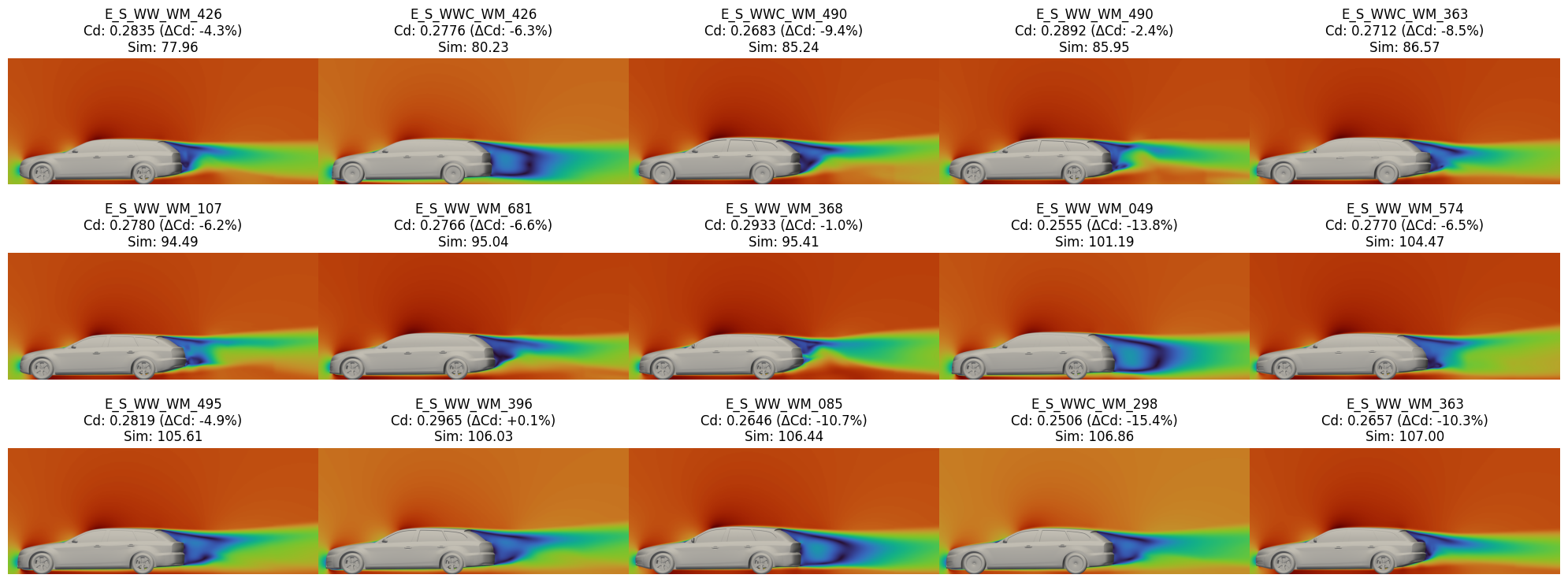}
\caption{Illustration of design retrieval from DrivAerNet++ using either geometric shape similarity or aerodynamic performance metrics. Given a query design, similar cars are identified based on shape or aerodynamic performance characteristics, enabling efficient exploration and optimization of car designs. It also quantifies the improvement in terms of aerodynamic drag by showing the difference in drag coefficient ($\Delta C_D$) with respect to a baseline design.}
\label{fig:design_retrieval}
\end{figure*}

\paragraph{Automatic Mesh Quality Verification and Optimization}
After each mesh generation, the Meshing Agent automatically executes OpenFOAM's \textit{checkMesh} utility to assess mesh quality from both geometric and topological perspectives, as summarized in Table~\ref{tab:mesh_quality}. While passing these checks does not guarantee simulation convergence, it is an essential step to ensure suitable geometry and topology characteristics, significantly contributing to simulation stability and accuracy.

\begin{table}[h!]
    \footnotesize
    \centering
    \renewcommand{\arraystretch}{1.3}
    \begin{tabular}{lccc}
        \hline
        \textbf{Quality Test} & \textbf{Mesh 1} & \textbf{Mesh 2} & \textbf{Mesh 3} \\
        \hline
        Boundary Openness & \textcolor{mydarkgreen}{\CheckmarkBold} & \textcolor{mydarkgreen}{\CheckmarkBold} & \textcolor{mydarkgreen}{\CheckmarkBold} \\
        Max Aspect Ratio  & \textcolor{mydarkgreen}{\CheckmarkBold} &  \textcolor{mydarkgreen}{\CheckmarkBold} & \textcolor{mydarkgreen}{\CheckmarkBold} \\
        Min Face Area     & \textcolor{mydarkgreen}{\CheckmarkBold} & \textcolor{mydarkgreen}{\CheckmarkBold} & \textcolor{mydarkgreen}{\CheckmarkBold} \\
        Max Face Area     & \textcolor{mydarkgreen}{\CheckmarkBold} & \textcolor{mydarkgreen}{\CheckmarkBold} & \textcolor{mydarkgreen}{\CheckmarkBold} \\
        Min Volume        & \textcolor{mydarkgreen}{\CheckmarkBold} & \textcolor{mydarkgreen}{\CheckmarkBold} &  \textcolor{mydarkgreen}{\CheckmarkBold} \\
        Max Volume        &  \textcolor{mydarkgreen}{\CheckmarkBold} & \textcolor{mydarkgreen}{\CheckmarkBold} & \textcolor{mydarkgreen}{\CheckmarkBold} \\
        Mesh Non-Orthogonality & \textcolor{mydarkgreen}{\CheckmarkBold} & \textcolor{mydarkgreen}{\CheckmarkBold} & \textcolor{mydarkgreen}{\CheckmarkBold} \\
        Face Pyramids     & \textcolor{mydarkgreen}{\CheckmarkBold} & \textcolor{mydarkgreen}{\CheckmarkBold} & \textcolor{mydarkgreen}{\CheckmarkBold} \\
        Max Skewness      & \textcolor{mydarkgreen}{\CheckmarkBold} & \textcolor{mydarkgreen}{\CheckmarkBold} & \textcolor{mydarkgreen}{\CheckmarkBold} \\
        Coupled Point Match & \textcolor{mydarkgreen}{\CheckmarkBold} & \textcolor{mydarkgreen}{\CheckmarkBold} & \textcolor{mydarkgreen}{\CheckmarkBold} \\
        \hline
    \end{tabular}
        \vspace{2pt}
    \caption{Mesh quality test results for three different meshes. \textcolor{mydarkgreen}{\CheckmarkBold} indicates the test was passed, while \textcolor{mydarkred}{\XSolidBrush} indicates failure.}
    \label{tab:mesh_quality}
\end{table}

\subsection{Real-time Aerodynamic Analysis with the Simulation Agent}

The \textit{Simulation Agent} enables real-time aerodynamic performance assessment and design retrieval using the DrivAerNet++ dataset. Given a 2D hand-drawn sketch or a 3D design, it efficiently searches through the database of 8,000 high-fidelity CFD simulations, retrieving designs based on geometric shape similarity or aerodynamic performance metrics. Additionally, it leverages surrogate models to rapidly predict aerodynamic properties such as drag coefficient and flow patterns, allowing designers to immediately quantify potential performance improvements and analyze the associated flow characteristics. The results from the Simulation Agent support design exploration and optimization tasks by offering instant insights into aerodynamic performance, streamlining the iterative design process, and enhancing decision-making.

\subsubsection{Sketch-Based Rapid Retrieval of Aerodynamic Simulation Results}

The Simulation Agent facilitates real-time retrieval of aerodynamic performance data based on 2D hand-drawn sketches or design inputs. Given a sketch or conceptual design, it queries the DrivAerNet++ database containing 8,000 high-fidelity CFD simulations in real-time to rapidly identify and retrieve aerodynamic properties such as drag coefficient, pressure distribution, and velocity fields. 

We use the same approach employed for the Styling Agent, utilizing a pre-trained CNN, specifically ResNet50~\cite{he2016deep}, to evaluate design diversity. Given a set of $N$ images, each image \( i \) is represented by a feature vector \( \mathbf{f}_i \in \mathbb{R}^d \). We then calculate the diversity score by computing the average pairwise L2 distance between these feature vectors. Images exhibiting higher average distances are considered more diverse, reflecting greater visual or stylistic variance within the dataset. As shown in Figure~\ref{fig:design_retrieval}, the retrieval system identifies similar designs and provides direct aerodynamic comparisons. Engineers can quantify the improvements in aerodynamic drag and analyze the wake flow, gaining insights into flow separation and turbulence characteristics. This process bridges the gap between early-stage conceptual design and performance-driven engineering, enabling data-driven decisions for optimizing aerodynamics while maintaining stylistic intent.

Since DrivAerNet++ includes multiple modalities for each design, such as parametric data, point clouds, and part annotations, this additional information can be leveraged for various downstream tasks. For instance, parametric data can be utilized for manufacturing processes and constraints, while point cloud representations enable advanced shape analysis and reverse engineering. Furthermore, part annotations provide structured insights into component-level performance, facilitating targeted design optimizations and generative modeling for aerodynamic and structural improvements.

\subsubsection{Real-time Aerodynamic Predictions with Surrogate Modeling}

In this work, we utilize TripNet~\cite{chen2025tripnet}, a triplane neural architecture, within the Simulation Agent to rapidly and accurately predict aerodynamic properties directly from 3D car geometries. This enables the Simulation Agent to predict aerodynamic properties such as drag coefficients, surface pressure distributions, and full volumetric flow fields in real-time, significantly accelerating aerodynamic analyses. The model is trained on the DrivAerNet++ dataset and evaluated on an unseen test set consisting of 1,200 industry-standard car designs, providing a comprehensive evaluation due to the extensive size of the test set.

\begin{figure}[h!]

    \includegraphics[width=\linewidth]{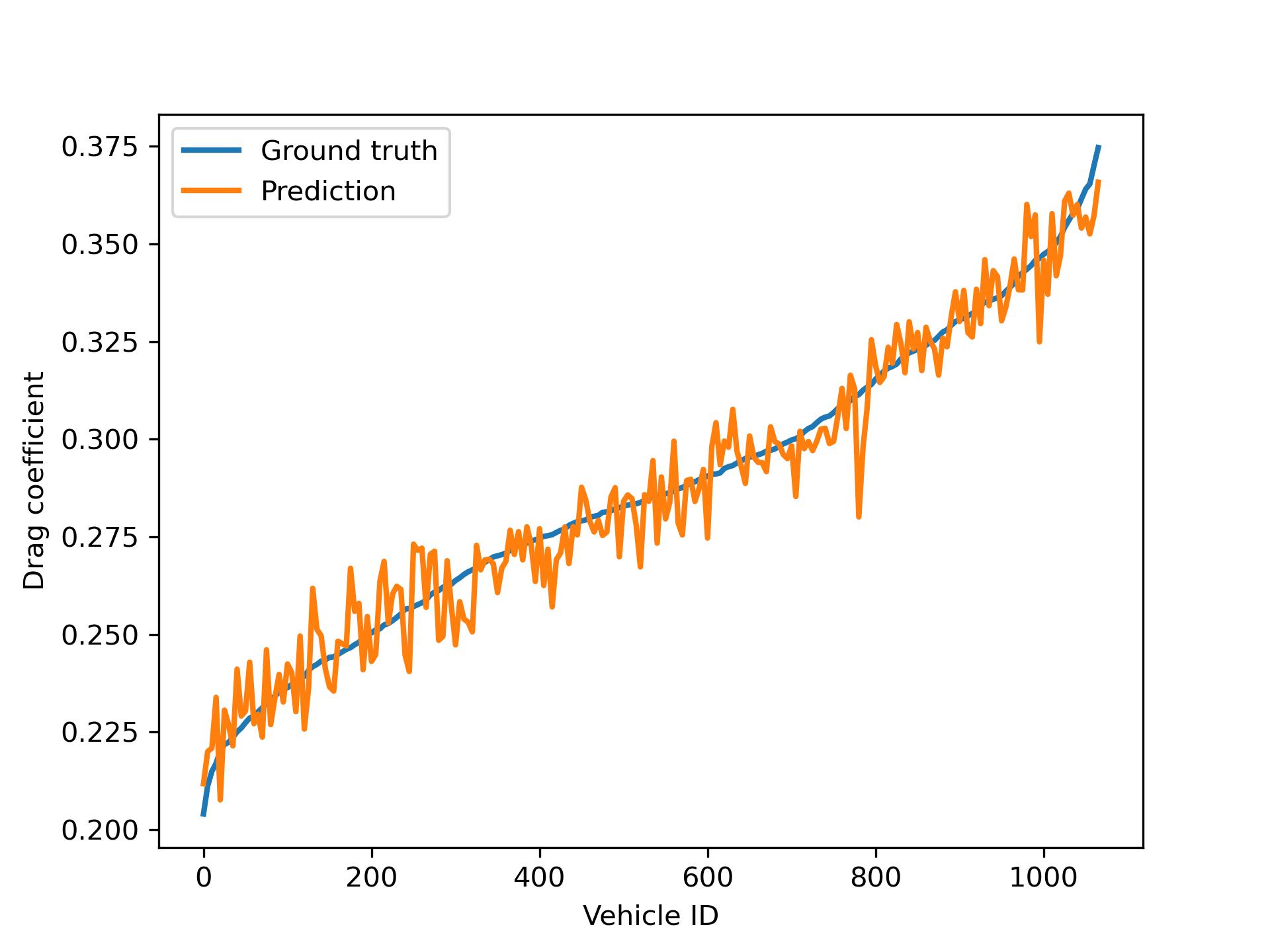}
    \caption{Comparison of ground truth and predicted drag coefficient trends across different car designs from the DrivAerNet++ unseen test set, which includes 1200 industry-standard car designs. The model effectively captures the overall trend, with acceptable oscillatory behavior in the predictions.}
    \label{fig:drag_trend}
\end{figure}

Figure~\ref{fig:drag_trend} illustrates the design trends in drag coefficient and a comparison between ground truth and predicted values for various car designs. The ground truth values are arranged in ascending order of drag coefficient and plotted alongside the predicted values for the same designs. The model effectively captures the overall trend, with predictions closely following the ground truth. However, the predicted values exhibit some oscillatory behavior, particularly in smaller directional variations between successive designs. This oscillation is acceptable and does not significantly impact the overall trend, as the model effectively captures the general pattern of drag coefficient variation. 

\begin{figure}[h!]
    \includegraphics[width=\linewidth]{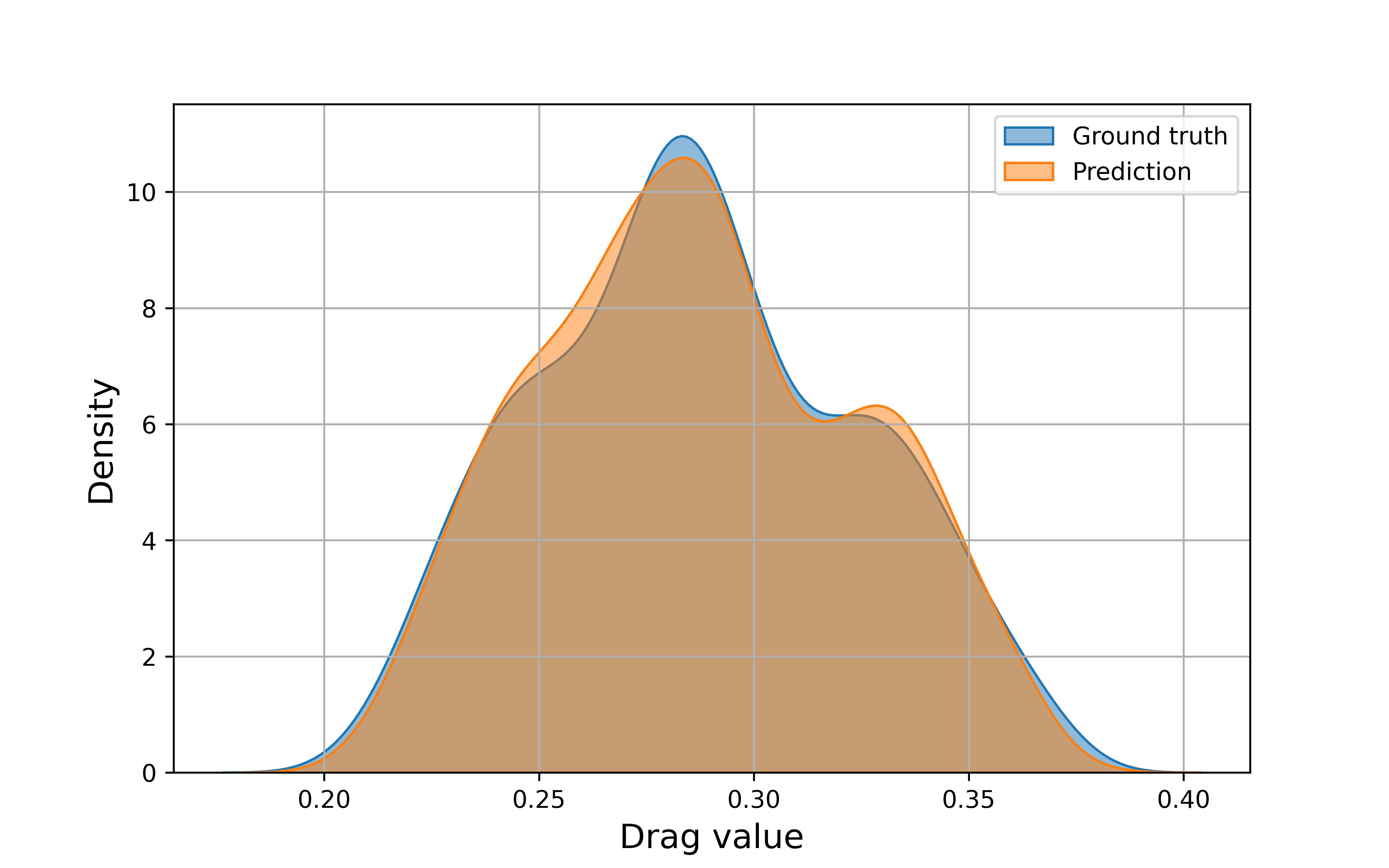}
    \caption{Probability density distribution of drag coefficients for ground truth (CFD simulations) and model predictions. The strong overlap indicates that the model accurately replicates the statistical distribution of the aerodynamic performance of the unseen test set.}
    \label{fig:kde}
\end{figure}

Figure~\ref{fig:kde} presents the probability density distribution of drag coefficient values for both the ground truth from CFD simulations and the model predictions. The distributions closely align, demonstrating that the predicted drag coefficients follow a similar statistical pattern to the ground truth values. The strong overlap between the two distributions indicates that the model effectively captures the overall variability and range of aerodynamic performance across different car designs. Minor deviations in the predicted distribution, such as slight differences in peak density and secondary modes, suggest areas for potential refinement but do not significantly impact the overall agreement. This result highlights the model’s capability in producing realistic aerodynamic predictions, reinforcing its applicability in engineering simulations.

Finally, the results in Table~\ref{tab:drag_transition} demonstrate the simulation agent’s capability to accurately predict differences in drag coefficients ($\Delta C_D$) between various car configurations compared to CFD simulations. For instance, the predicted drag difference between the FD-FS and FD-N configurations is 0.066, closely matching the CFD value of 0.065. Similarly, the simulation agent correctly predicts the drag difference between the E-FS and E-N configurations, with only minor deviations of 0.001. The largest discrepancy appears in the N-FS configuration, where the predicted value is -0.00007 compared to the CFD result of -0.0003, though this remains within an acceptable range. 

\begin{table}[h!]
    \centering
    \begin{tabular}{lcc}
    \hline
    & CFD Simulation & Simulation Agent \\
    \hline
    FD-FS & 0.065 & 0.066\\ 
    FD-N & 0.065 & 0.066 \\ 
    E - FD  &  $-$0.037 &  $-$0.038 \\
    N - FS  & $-$0.0003 &  $-$0.00007 \\
    E - FS  & 0.028 & 0.028  \\
    E - N  & 0.029 &  0.028 \\
    \hline
    \end{tabular}
        \vspace{2pt}
        \caption{Comparison of differences in drag coefficients ($\Delta C_D$) from CFD simulations and the Simulation Agent (TripNet). Results are computed on an unseen test set comprising 1,200 diverse car designs from the DrivAerNet++ dataset, covering estateback (E), notchback (N), fastback smooth underbody (FS), and fastback detailed underbody (FD) configurations.}

    \label{tab:drag_transition}
\end{table}


\section{Results and Observations}
\label{sec:results}

Our method enables AI-assisted rapid iteration in car design by generating aesthetic variations while preserving the underlying structural integrity of the cars. The application of LLMs and VLMs in stylization introduces creative diversity, making this technique suitable for industrial applications such as automotive concept design and custom styling.

To generate high-quality stylized car designs, we utilize Stable Diffusion XL (SDXL)~\cite{podell2023sdxl}, a latent diffusion model capable of producing photorealistic and artistically enhanced images based on text prompts. SDXL leverages a dual-stage architecture with a base model for coarse generation and a refinement model for high-detail enhancements. The model is conditioned using ControlNet, allowing us to guide the generation process with Canny edge maps, ensuring that the output images maintain the structural integrity of the original car designs. This approach allows for precise control over styling variations, enabling the synthesis of diverse and visually appealing automotive concepts.

The results primarily summarize the workflow outlined in Section~\ref{sec:DesignAgents}. Beginning with a 2D sketch of an estateback car design, the Styling Agent generated high-resolution rendered images, as shown in Figure~\ref{fig:car_styling_sdxl}. Next, the CAD Agent retrieved the most similar designs from the DrivAerNet++ database and further generated new 3D car shapes by interpolating between the estateback design and three other configurations (Figure~\ref{fig:deep_sdf_interpolation}). The Meshing Agent then processed the generated 3D models, producing high-quality CFD meshes optimized for aerodynamic simulations. Finally, the Simulation Agent performed real-time aerodynamic analysis by retrieving the most similar designs from the DrivAerNet++ database and obtaining corresponding aerodynamic performance data, as illustrated in Figure~\ref{fig:design_retrieval}.

\section{Implications for Design}
\label{sec:implications}
The integration of AI-driven Design Agents into automotive design is transforming the traditional workflow by automating key tasks such as concept generation, performance evaluation, and design optimization. Our multi-agent system leverages generative models, geometric deep learning, and high-fidelity simulations to enable engineers and designers to explore a vast design space efficiently. By orchestrating interactions between specialized agents—ranging from styling and CAD retrieval to meshing and simulation—the framework accelerates the iterative design process while maintaining industry-standard engineering constraints.

As these AI-powered agents continue to evolve, their ability to collaborate with human designers will redefine the future of automotive engineering. Rather than replacing human intuition, these agents serve as intelligent assistants, augmenting creativity and providing real-time feedback on aerodynamic performance and manufacturability. This approach not only streamlines the development pipeline but also ensures that novel designs are both aesthetically compelling and functionally optimized. While our framework specifically addresses automotive design, the implications of AI-driven design agents extend broadly to any engineering domain involving performance-driven simulations, such as aerospace, naval architecture, or structural engineering. Ultimately, this approach fosters a more interactive and iterative design environment, bridging creativity, aesthetics, and functional requirements, and reshaping traditional design methodologies by significantly enhancing efficiency and creativity through collaborative human-AI workflows.

\section{Limitations, Future Work, and Conclusion}
\label{sec:conclusion}
One limitation of our approach is that mesh evaluation currently focuses primarily on geometric and topological quality metrics obtained from utilities like \textit{checkMesh}. However, accurate mesh evaluation should also incorporate validation through CFD simulation outcomes, experimental data, and adherence to established best practices. Additionally, aesthetics evaluation can be inherently subjective; thus, future work should include surveys with engineers, designers, and potential users to quantitatively assess the visual appeal and practicality of generated designs.

Another avenue for future work involves exploring alternative multi-agent orchestration frameworks beyond AutoGen to assess whether different systems could further enhance agent collaboration, scalability, or task management effectiveness. Finally, while our approach focuses on early-stage conceptual design, styling, and aerodynamic evaluation, future work could extend these capabilities to prototyping, manufacturing, and market-driven design analysis.

Finally, we conclude by presenting an AI-driven multi-agent framework leveraging state-of-the-art vision-language models (VLMs), large language models (LLMs), and geometric deep learning methods to support and accelerate key stages of the automotive design process. Our approach uniquely integrates aesthetics and aerodynamic performance using a high-fidelity, large-scale dataset comprising 8,000 industry-standard designs. We demonstrated how this multi-agent system facilitates seamless interaction among designers, engineers, deep learning and generative AI models, as well as engineering software such as OpenFOAM, ParaView, and Blender. The framework enables real-time retrieval of CAD models, parametric representations, or CFD simulation data based on geometric and performance similarity. Additionally, intuitive text prompts streamline styling and meshing tasks, allowing newly generated designs to be rapidly meshed or evaluated for aerodynamic performance through surrogate models. These capabilities highlight the transformative potential of integrated AI-driven methods in engineering design workflows, encouraging further research and practical adoption across academia and industry.





\bibliographystyle{asmeconf}  
\bibliography{asmeconf-sample}

\appendix


\end{document}